\pdfoutput=1

\documentclass[11pt]{article}

\usepackage[preprint]{acl}

\usepackage{times}
\usepackage{latexsym}
\usepackage{graphicx}
\usepackage{array}
\usepackage{makecell}
\usepackage{bm}
\usepackage{multirow}
\usepackage{amsfonts}
\usepackage{xcolor} 
\usepackage{colortbl} 
\usepackage{booktabs}
\usepackage{url}

\usepackage{hyperref}

\usepackage[T1]{fontenc}

\usepackage[utf8]{inputenc}

\usepackage{microtype}

\usepackage{inconsolata}
\newcommand\blfootnote[1]{%
  \begingroup
  \renewcommand\thefootnote{}\footnote{#1}%
  \addtocounter{footnote}{-1}%
  \endgroup
}

\title{Multimodal Table Understanding}

\author{Mingyu Zheng$^{1,2*\dagger}$,\ Xinwei Feng$^{3*}$,\ Qingyi Si$^{1,2*}$,\ Qiaoqiao She$^{3}$, \\ {\bf Zheng Lin$^{1,2\ddagger}$, Wenbin Jiang$^4$, Weiping Wang$^{1}$}  \\
$^1$Institute of Information Engineering, Chinese Academy of Sciences, Beijing, China \\
$^2$School of Cyber Security, University of Chinese Academy of Sciences, Beijing, China \\
$^3$Baidu Inc, Beijing, China \\
$^4$School of Artificial Intelligence, Beijing Normal University, Beijing, China \\
\texttt{\{zhengmingyu,siqingyi,linzheng,wangweiping\}@iie.ac.cn}\\
\texttt{\{fengxinwei,sheqiaoqiao\}@baidu.com}\\
\texttt{\{jiangwenbin\}@bnu.edu.cn}\\
}

\begin{document}
\maketitle
\begin{abstract}

Although great progress has been made by previous table understanding methods including recent approaches based on large language models (LLMs), they rely heavily on the premise that given tables must be converted into a certain text sequence (such as Markdown or HTML) to serve as model input. However, it is difficult to access such high-quality textual table representations in some real-world scenarios, and table images are much more accessible. Therefore, how to directly understand tables using intuitive visual information is a crucial and urgent challenge for developing more practical applications. In this paper, we propose a new problem, multimodal table understanding, where the model needs to generate correct responses to various table-related requests based on the given table image. To facilitate both the model training and evaluation, we construct a large-scale dataset named MMTab, which covers a wide spectrum of table images, instructions and tasks. On this basis, we develop Table-LLaVA, a generalist tabular multimodal large language model (MLLM), which significantly outperforms recent open-source MLLM baselines on 23 benchmarks under held-in and held-out settings. The code and data is available at \url{https://github.com/SpursGoZmy/Table-LLaVA}.\blfootnote{$^{*}$ Indicates equal contribution.}
\blfootnote{$^{\dagger}$This work was done during an internship at Baidu Inc.}
\blfootnote{$^{\ddagger}$ Corresponding author: Zheng Lin.}


\end{abstract}

\section{Introduction}
Tables are widely used to store and present data across various fields, e.g., financial analysis, scientific research and government reports~\citep{web_table_taxonomy,tu_survey_2023}. To make the most of the abundant tabular data, the table understanding (TU) technique has been proposed to automatically understand tables and perform table-based tasks, such as question answering~\citep{WTQ} and text generation~\citep{ToTTo}. As a technique that could significantly elevate work efficiency in different industries, it has attracted ever-increasing research interest in recent years.

\begin{figure}[t]
  \centering
  \includegraphics[width=0.82\linewidth]{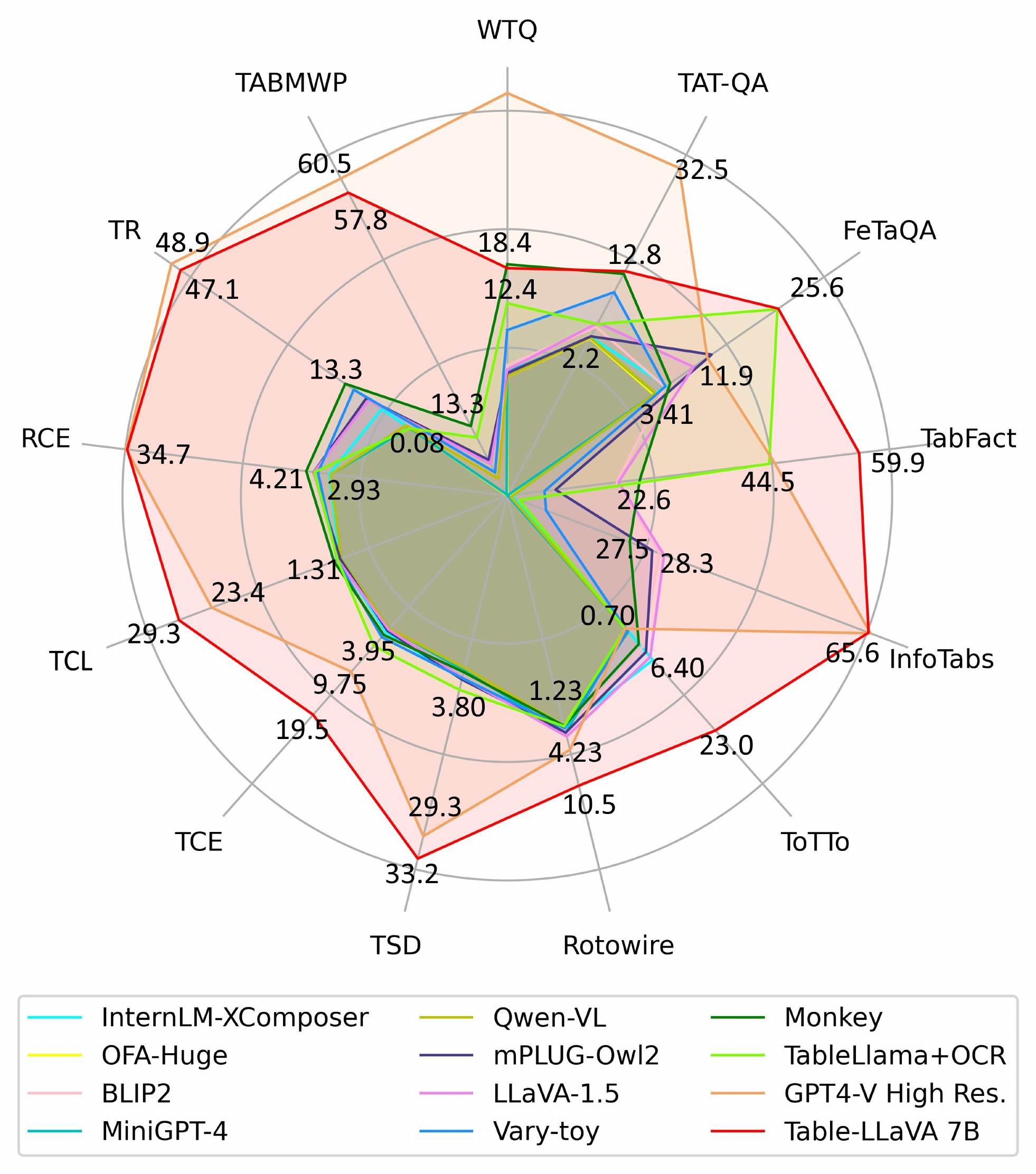}
  \caption{An overall performance comparison of Table-LLaVA 7B and existing MLLMs on various multimodal table understanding benchmarks. Table-LLaVA outperforms recent open-source MLLMs and is even competitive with the powerful GPT-4V on most tasks.}
  \label{radar_graph}
  
\end{figure}

Though considerable efforts have been dedicated to the table understanding problem~\citep{TAPAS,liu2022tapex}, most previous models can only fulfill very limited tasks until the emergence of large language models (LLMs)~\citep{gpt-3,palm}. With the help of powerful LLMs, we are getting closer to the vision that a versatile model can perform a variety of table-based tasks. However, existing table-oriented LLMs~\citep{tablellama,tablegpt_chatgpt, tablegpt_phoenix} rely heavily on the prerequisite that all given tables must be converted into a certain text sequence (like Markdown or HTML) to be input to LLMs. Under some practical scenarios like scanned documents and webpage screenshots, it is difficult to obtain such high-quality textual table representations, and yet table images are more accessible. Moreover, 
humans can directly understand two-dimensional tables using the intuitive visual information, whereas LLMs can only interpret tables in a one-directional textual perspective, which may increase the difficulty of comprehending diverse table structures and colored table elements. In summary, for the sake of convenience and intuitiveness, it is a crucial and urgent challenge to explore how to directly digest tables using visual information.

To promote the advancement of table understanding and its real-world applications, we propose the \textbf{multimodal table understanding} problem, where the model is required to generate correct responses to different table-related requests (e.g., questions) in an end-to-end fashion based on the table image. Despite the fact that recent multimodal large language models (MLLMs) have demonstrated excellent capabilities in many multimodal tasks, they cannot be directly extended to the proposed task. As shown in Figure \ref{radar_graph}, the performance of popular MLLMs like MiniGPT-4~\citep{minigpt4} and BLIP2~\citep{blip2} is close to zero on most tasks, revealing their weakness in understanding tabular data. More importantly, 
there is a lack of a comprehensive dataset that can support both the development and evaluation of generalist MLLMs towards multimodal table understanding. 

To address the above issue, we construct \textbf{MMTab}, the first open-source large-scale dataset for multimodal table understanding problem, based on 14 publicly available table datasets of 8 domains. We carefully design scripts to convert original textual tables in these datasets into table images highlighting a broad coverage of table structures and styles, and transform all task-specific samples into multimodal instruction-tuning samples with a unified format of \texttt{<table image, input request, output response>}. The resulting dataset contains (1) 150K table recognition samples on 97K table images for pre-training (named \textbf{MMTab-pre}). (2) 232K samples of 14 table-based tasks on 82K table images for instruction tuning (named \textbf{MMTab-instruct}). (3) 49K test samples on 23K table images composing 17 held-in and 7 held-out benchmarks (named \textbf{MMTab-eval}). During the dataset construction, data augmentations at multiple levels (e.g., table-level, task-level) were adopted to further improve the data diversity, and we also introduce multimodal table structure understanding tasks that have been overlooked in previous studies. 

Based on the curated dataset, we develop a versatile tabular MLLM named \textbf{Table-LLaVA} with an enhanced two-stage training paradigm. In the first stage, we pre-train LLaVA-1.5~\citep{improved_llava} with an extra table recognition task on the MMTab-pre, which requires the model to generate textual sequences (like HTML) given table images. This stage aligns the structures and elements within table images to textual modality and thus enhances the comprehension of the basic table structure and content. In the second stage, we continue to instruction-tuning the model with diverse table-based downstream tasks on the MMTab-instruct, which endows the model with the multimodal instruction-following ability for table-related requests.

We compare Table-LLaVA with a series of open-source (M)LLMs and closed-source GPT-4V. Experimental results show that Table-LLaVA beats strong MLLM baselines on 17 held-in and 6 held-out benchmarks, and is even competitive with the powerful GPT-4V on 14 benchmarks with a subset of test samples. Extensive ablation experiments are conducted to reveal the contributions of different training data (e.g., the influence of table recognition pre-training data). We also explore the mutual influence between model's capacity for tabular tasks and non-tabular tasks. We hope this work could establish a strong base for future research on the multimodal table understanding problem and facilitate the progress of more general MLLMs.

We conclude our contributions as follows:

1) We make the first systematic exploration of the multimodal table understanding problem, which is complementary to the traditional text-only problem setting. 

2) Accordingly, we construct and release a large-scale dataset MM-Tab which covers diverse tables and data for different tasks, including a series of novel table structure understanding tasks.

3) We develop a versatile tabular MLLM Table-LLaVA,  which significantly outperforms a range of strong MLLM baselines under both held-in and held-out settings (Figure \ref{radar_graph}).

\section{Related Work}
\subsection{Table Understanding}

The table understanding (TU) problem concentrates on how to automatically extract, transform and interpret essential information from tabular data, and it has attracted significant attention in the past years~\citep{tu_survey_2021,tu_survey_2023}. Many tasks fall under the umbrella of table understanding problem, e.g., Table Question Answering (TQA)~\citep{FeTaQA,IM-TQA}, Table Fact Verification (TFV)~\citep{TabFact} and Table-to-Text (T2T) generation~\citep{HiTab}. 

Different approaches have been proposed to solve specific TU tasks, ranging from early rule-based systems~\citep{ventex} to later tabular language models (TaLMs)~\citep{liu2022tapex,hytrel,table_pretraining_survey}, which are pre-trained from general language models like BERT~\citep{devlin2019bert} with extra large-scale table corpus. Nevertheless, these methods can only support limited TU tasks and handle tables of specific types. Recently, the emerging LLMs have opened up new possibilities for utilizing one single model to fulfill multiple table tasks. Researchers have attempted to enhance the TU ability of existing LLMs with different strategies such as prompt engineering~\citep{table_cot,gpt4table}, instruction tuning~\citep{tablellama,tablegpt_chatgpt,liu2023zero} and combining external tools~\citep{lu2023chameleon,sheetcopilot}. The resulting tabular LLMs like TableLlama~\citep{tablellama} and TableGPT~\citep{tablegpt_chatgpt} can possess better TU ability and respond to wide-ranging table-related requests. However, previous TU approaches including tabular LLMs are unable to directly understand table images, which limits the potential application scenarios of TU technique.

\subsection{Multimodal Large Language Models}

With LLMs experiencing rapid advancements, recent studies have tried to endow the purely texutal LLMs with understanding and perception capabilities of other modalities such as image and video, leading to the emergence of MLLMs~\citep{alayrac2022flamingo,li2022blip_model}. Flamingo~\citep{alayrac2022flamingo} proposes a gated cross-attention mechanism between vision encoder and LLM, which is trained on billions of image-text pairs to align vision and language modalities. BLIP2~\citep{blip2} introduces a Q-Former with learnable query vectors to abstract the visual information from vision encoder into features of a fixed number. LLaVA~\citep{liu2023llava_1} uses a linear layer as a simpler cross-modal connector and achieve powerful performance with better data efficiency.

Though previous MLLMs demonstrated remarkable performance on multiple multimodal tasks~\citep{liu2023mmbench,yu2023mmvet}, their ability to digest table images and perform downstream tasks has not been thoroughly investigated. In this work, we build the first large-scale multimodal table understanding dataset and develop Table-LLaVA, a versatile tabular MLLM for diverse table-based tasks. To stimulate future endeavours on this problem, we also provide a comprehensive benchmark and fully evaluate the table understanding ability of existing models. More recently, researchers also tried to develop MLLMs like Vary~\citep{wei2023vary} and Monkey~\citep{li2023monkey} to understand document pictures with enhanced visual encoders, e.g., scaling up the vision vocabulary and image resolution. These models focus on the unified visual understanding of different document images and can be further improved with the proposed dataset.

\begin{figure*}[t]
  \centering
  \includegraphics[width=\linewidth]{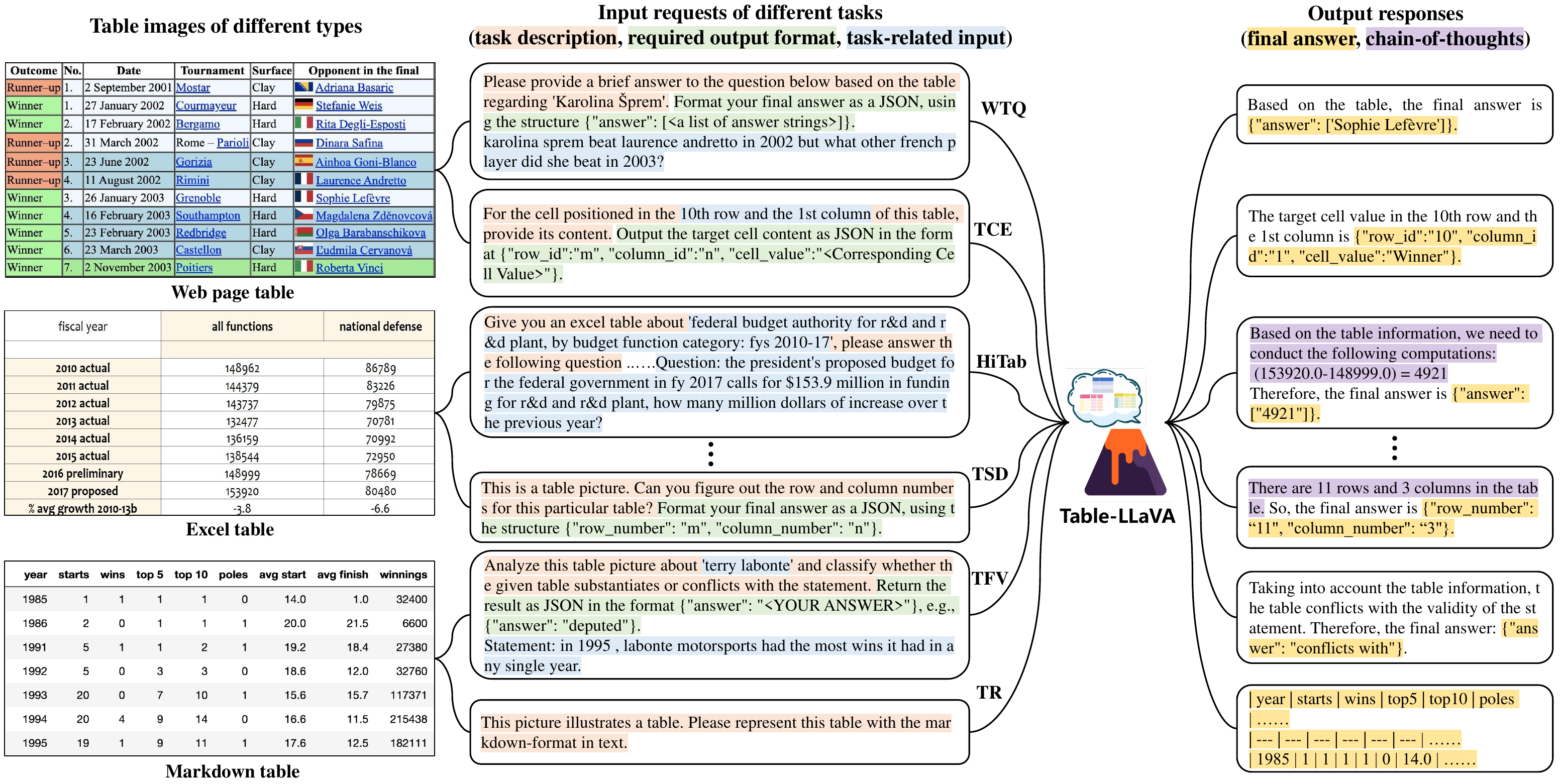}
  \caption{MMTab contains diversified table images and instruction following data, covering a wide range of tabular tasks (see Table \ref{dataset_statistics}). More dataset examples are shown in Figure \ref{more_dataset_example_1}-\ref{more_dataset_example_2} in Appendix \ref{more_dataset_examples}.
  }
  \label{dataset_example}
\end{figure*}

\section{MMTab Dataset}

\begin{figure}[t]
  \centering
  \includegraphics[width=\linewidth]{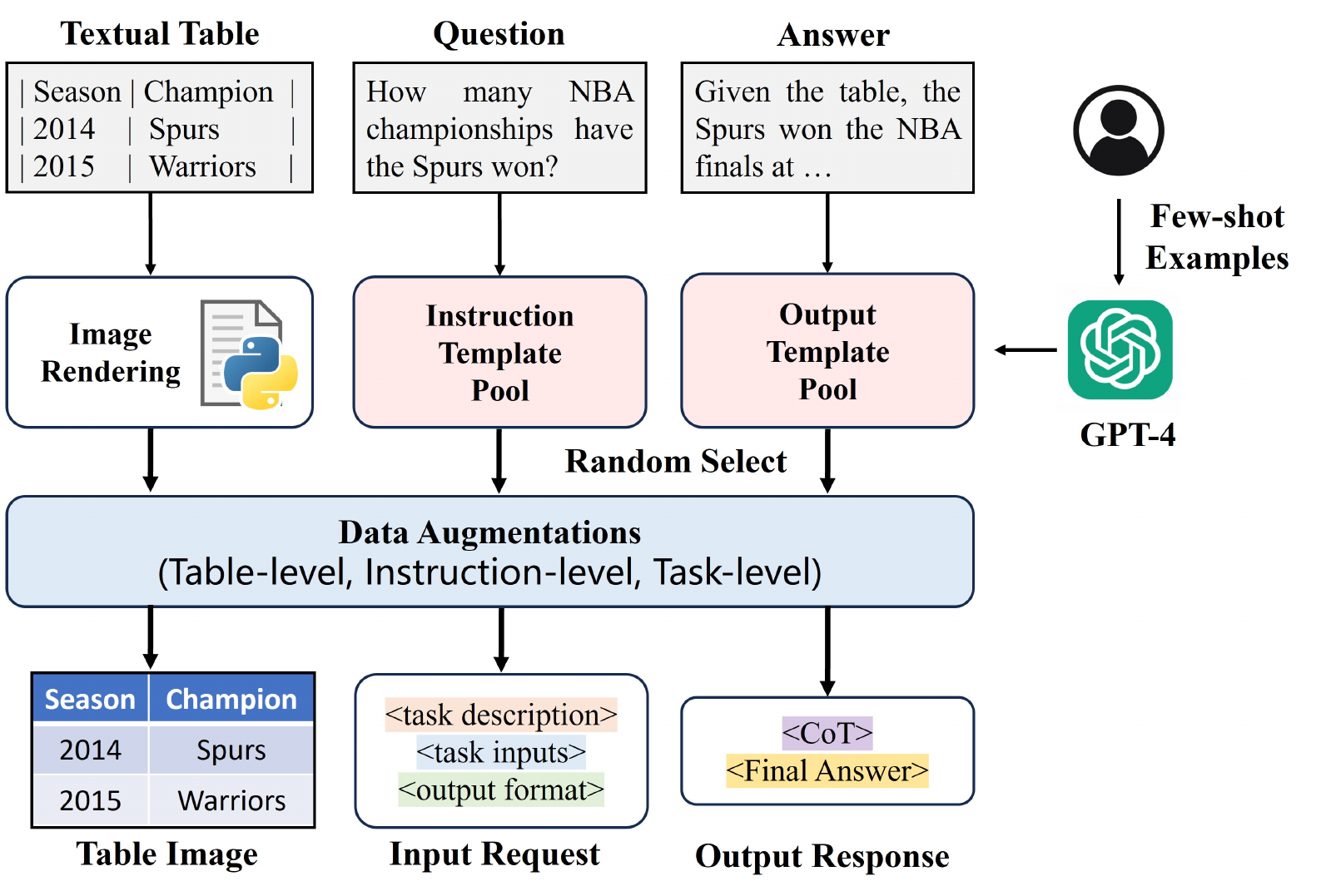}
  \caption{The overview of sample construction process.}
  \label{sample_collection_overview}
  
\end{figure}

\subsection{Data Collection}
\label{data_collection}

As shown in Table \ref{dataset_statistics}, with a pursuit of diverse table structures, tasks, and domains, we collect samples from 14 public table datasets of 8 domains (the first 14 rows in Table \ref{dataset_statistics}), covering 9 representative academic tasks. The detailed definition of each task can be found in Table \ref{task_definitions}. The original tables in these datasets are stored in divergent textual formats such as HTML or Markdown. We carefully design Python scripts with external packages like html2image to convert textual tables into high-quality table images. The task-specific input and output texts are transformed into the instruction-following format with pre-defined instruction templates. To minimize errors during answering parsing, we also add extra instructions, requiring models to output the final answer in the JSON format. As shown in the Figure \ref{dataset_example}, the rendered table images and processed input-output pairs constitute the final multimodal instruction-tuning samples with a unified format of \texttt{<table image, input request, output response>}. We adhere to the original dataset partitioning and select 11 datasets for training and held-in evaluation. 3 small-scale datasets with non-overlapping domains are used for held-out evaluation. The overview of sample construction process is depicted in Figure \ref{sample_collection_overview}.

\subsection{Data Augmentations}
\label{data_augmentations}
Previous works have shown that the diversity of instruction-following data is crucial to the capability of the resulting instruction-following models~\citep{zhou2023lima,alpaca_cot,tablegpt_chatgpt}. To create more data diversity and avoid over-fitting in the model training, we perform additional data augmentations at multiple levels.

\textbf{Table-level augmentations.} Real-world tables often have varied structures and styles. An ideal table understanding model should be able to process divergent tables like a human reader. Since our dataset already includes diverse table structures from academic datasets, we separately design scripts to render table images with three different styles: Web-page (70.8\%), Excel (19.4\%) and Markdown (9.8\%). Fine-grained adjustments such as font type and cell colors are also considered. 

\textbf{Instruction-level augmentations.} In practical scenarios, user instructions for the same task are likely to vary from user to user. To improve models' robustness towards such variations, we resort to GPT-4 to generate new instruction templates and descriptions about JSON output format in a few-shot fashion based on several manually annotated demonstrations. Generated instruction templates with grammar mistakes or deviation from the original task are filtered out. When we construct input requests of each dataset, we randomly select an instruction template and an output format description from the candidate pool, and then combine them with the task-specific input such as table-related questions to produce the final input request. This combination strategy can bring more diversity of input requests. Using the TABMWP as an example, we show instruction templates and Python code for building input requests in Figure \ref{instruction_building_process}.

\textbf{Task-level augmentations.} Though the collected 14 public datasets highlight 9 academic tabular tasks (e.g., Flat TQA and Cell Description) which demand table-based reasoning capabilities, it is still a question whether existing MLLMs are truly aware of the basic table structures. Prior study has found that, despite achieving great performance on downstream tasks, tabular LLMs may still exhibit poor capacity for perceiving table structures~\citep{gpt4table}. To further strengthen the fundamental table structure understanding ability of MLLMs, 6 table structure understanding tasks (the 6 rows with `Structure Understanding' task category in Table \ref{dataset_statistics}) are devised, e.g., table size detection (TSD) task. For each task, we use the above-mentioned method to generate input requests and design scripts to automatically extract the final answer from the textual tables in collected datasets. Finally, 8K training samples, 1K or 1.25K evaluation samples were constructed for each structure understanding task. Except above strategies, we also combine single-turn samples of the same table to compose 37K multi-turn conversation samples. At last, we obtain 232K samples on 82K table images for instruction-tuning (named \textbf{MMTab-instruct}), and 45K held-in and 4K held-out test samples on 23K table images for evaluation (named \textbf{MMTab-eval}).

Inspired by existing MLLMs which align textual descriptions with input images through image-text pre-training, we introduce the table recognition task as an important pre-training task for multimodal table understanding. In this task, MLLMs learn to generate a textual table representation such as an HTML sequence given the table image, which helps aligning structure and text information in the table image with the ground-truth. We additionally collect 20K table images from the ToTTo~\citep{ToTTo} training split and merge them with table images in the MMTab-instruct to construct sufficient pre-training data. Based on these table images and their original textual tables, we write scripts to construct table representations of three formats (HTML, Markdown and Latex), and then build instruction-following samples in the same way of MMTab-instruct. The resulting pre-training data contains 150K table recognition samples (HTML: 96K, Markdown: 27K, Latex: 27K) on 97K table images, which is denoted as \textbf{MMTab-pre}. More details about MMTab are given in Appendix \ref{more_details_of_dataset_construction}.

\begin{table*}[t]\footnotesize
\renewcommand{\arraystretch}{1.3}
\setlength\tabcolsep{2pt}
\centering
\scalebox{0.74}{
\begin{tabular}{c|c|c|cccc|cc|cc|c} 
\hline
\multirow{2}{*}{\textbf{MMTab}} & \multirow{2}{*}{\textbf{Task Category}} & \multirow{2}{*}{\textbf{Task Name}} & \multirow{2}{*}{\textbf{Dataset}} & \multirow{2}{*}{\textbf{Table Style}} & \multirow{2}{*}{\textbf{Domain}} & \multirow{2}{*}{\textbf{Held-in}} & \multicolumn{2}{c|}{\textbf{\# Tables}} & \multicolumn{2}{c|}{\textbf{\# Samples}} & \multirow{2}{*}{\begin{tabular}[c]{@{}c@{}}\textbf{Avg. Length}\\\textbf{(input/output)}\end{tabular}} \\ 
\cline{8-11}
 &  &  &  &  &  &  & Train & Test & Train & Test &  \\ 
\hline
\multirow{21}{*}{\begin{tabular}[c]{@{}c@{}}\textbf{MMTab-}\\\textbf{instruct}\end{tabular}} & \multirow{7}{*}{\begin{tabular}[c]{@{}c@{}}Table\\Question \\Answering\\(TQA)\end{tabular}} & Flat TQA & WTQ (\citeyear{WTQ}) & W & Wikipedia & Yes & 1.6K & 0.4K & 17K & 4K & 45.9/10.4 \\ 
\cline{3-12}
 &  & Free-form TQA & FeTaQA (\citeyear{FeTaQA}) & W & Wikipedia & Yes & 8K & 2K & 8K & 2K & 32.3/18.69 \\ 
\cline{3-12}
 &  & \multirow{2}{*}{Hierarchical TQA} & HiTab (\citeyear{HiTab}) & E & \begin{tabular}[c]{@{}c@{}}Wikipedia ~\\goverment reports\end{tabular} & Yes & 3K & 0.5K & 8K & 1.5K & 63.5/12.6 \\
 &  &  & AIT-QA (\citeyear{aitqa}) & E & Airline & No & - & 0.1K & - & 0.5K & 41.8/10.2 \\ 
\cline{3-12}
 &  & Multi-choice TQA & TabMCQ (\citeyear{tabmcq}) & M & science exams & No & - & 0.05K & - & 1K & 47.9/13.2 \\ 
\cline{3-12}
 &  & \multirow{2}{*}{\begin{tabular}[c]{@{}c@{}}Tabular\\Numerical Reasoning\end{tabular}} & TABMWP (\citeyear{tabmwp}) & W & math exams & Yes & 30K & 7K & 30K & 7K & 54.2/51.9 \\
 &  &  & TAT-QA (\citeyear{zhu2021tatqa}) & M & financial reports & Yes & 1.7K & 0.2K & 5.9K & 0.7K & 40.1/16.5 \\ 
\cline{2-12}
 & \multirow{3}{*}{\begin{tabular}[c]{@{}c@{}}Table Fact \\Verification (TFV)\end{tabular}} & \multirow{3}{*}{TFV} & TabFact (\citeyear{TabFact}) & E, M & Wikipedia & Yes & 9K & 1K & 31K & 6.8K & 49.9/18.3 \\
 &  &  & InfoTabs (\citeyear{infotabs}) & W & Wikipedia & Yes & 1.9K & 0.6K & 18K & 5.4K & 54.2/18.6 \\
 &  &  & PubHealthTab (\citeyear{pubhealthtab}) & W & public health & No & - & 0.3K & - & 1.9K & 71.9/18.4 \\ 
\cline{2-12}
 & \multirow{4}{*}{\begin{tabular}[c]{@{}c@{}}Table to\\Text \\(T2T)\end{tabular}} & \multirow{2}{*}{Cell Description} & ToTTo (\citeyear{ToTTo}) & W & Wikipedia & Yes & 15K & 7.7K & 15K & 7.7K & 31.1/14.8 \\ 
\cline{4-12}
 &  &  & HiTab\_T2T (\citeyear{HiTab}) & E & \begin{tabular}[c]{@{}c@{}}Wikipedia ~\\goverment reports\end{tabular} & Yes & 3K & 1.5K & 3K & 1.5K & 39.1/14.7 \\ 
\cline{3-12}
 &  & Game Summary & Rotowire (\citeyear{rotowire}) & E & NBA games & Yes & 3.4K & 0.3K & 3.4K & 0.3K & 27.6/291.7 \\ 
\cline{3-12}
 &  & Biography Generation & WikiBIO (\citeyear{wikibio}) & E & Wikipedia & Yes & 4.9K & 1K & 4.9K & 1K & 18.1/84.2 \\ 
\cline{2-12}
 & \multirow{6}{*}{\begin{tabular}[c]{@{}c@{}}Table\\Structure\\ Understanding\\(TSU)\end{tabular}} & Table Size Detection & TSD & W, E, M & - & Yes & 8K & 1.25K & 8K & 1.25K & 30.1/17.9 \\ 
\cline{3-12}
 &  & Table Cell Extraction & TCE & W, E, M & - & Yes & 8K & 1.25K & 8K & 1.25K & 51.6/19.9 \\ 
\cline{3-12}
 &  & Table Cell Locating & TCL & W, E, M & - & Yes & 8K & 1.25K & 8K & 1.25K & 72.5/45.6 \\ 
\cline{3-12}
 &  & Merged Cell Detection & MCD & W, E, M & - & Yes & 8K & 1K & 8K & 1K & 57.49/28.2 \\ 
\cline{3-12}
 &  & Row\&Column Extraction & RCE & W, E, M & - & Yes & 8K & 1.25K & 8K & 1.25K & 45.6/55.1 \\ 
\cline{3-12}
 &  & Table Recognition & TR & W, E, M & - & Yes & 8K & 1K & 8K & 1K & 16.3/389.2 \\ 
\cline{2-12}
 & \multicolumn{6}{c|}{Total} & 82K & - & 232K & - & 66.1/66.9 \\ 
\hline
\textbf{MMTab-eval} & \multicolumn{6}{c|}{Total} & - & 23K & - & 49K & 46.3/32.6 \\ 
\hline
\textbf{MMTab-pre} & \multicolumn{2}{c|}{Table Recognition} & \multicolumn{1}{c}{TR} & W, E, M & - & - & 97K & - & 150K & - & 16.4/397.5 \\
\hline
\end{tabular}
}
\caption{Breakdown statistics of the constructed \textbf{MMTab} dataset. W, E and M represents Web page, Excel, and Markdown tables, respectively. Task descriptions are shown in Table \ref{task_definitions} in Appendix \ref{more_dataset_examples}. For TSD, TCE, TCL, RCE datasets, their test samples contains 1K held-in and 0.25K held-out evaluation samples.}
\label{dataset_statistics}
\end{table*}

\subsection{Dataset Analysis}
\label{dataset_analysis}

\textbf{MMTab} offers the following advantages: \textit{(1) Large volume of data.} It contains 150K samples for pre-training, 232K samples for instruction-tuning, 45K samples and 4K samples for held-in and held-out evaluation, respectively. \textit{(2) Including tables of diverse structures, styles and domains.} It includes 105K table images covering a broad range of structures (e.g., simple tables with flat structures as well as complex tables with merged cells and hierarchical headers), divergent styles (i.e., Web page, Excel, and Markdown tables) and multiple domains (e.g., Wikipedia and financial reports). \textit{(3) Encompassing a wide range of tabular tasks.} In addition to 9 academic tasks which mainly evaluate the advanced table-based reasoning ability, MMTab also comprises 6 tasks aimed at assessing models' basic understanding of table structures. The broad coverage of tables and tasks can not only improve the generalization of the resulting model, but also provide a comprehensive testbed for MLLM research.

\section{Table-LLaVA}
\begin{figure}[t]
  \centering
  \includegraphics[width=\linewidth]{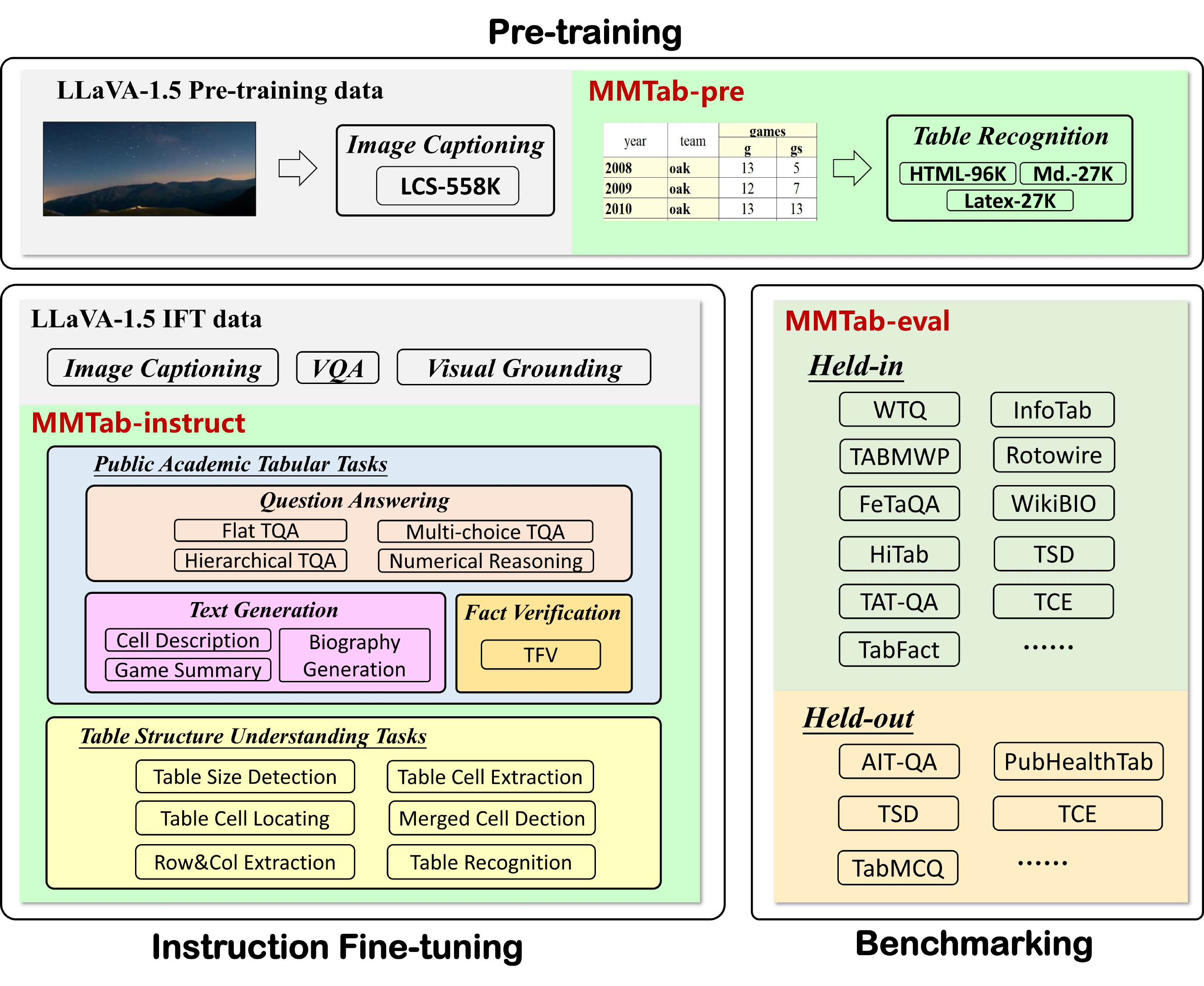}
  \caption{The two-stage training tasks and evaluation of Table-LLaVA. The red font represents our contribution.}
  \label{model_overview}
  
\end{figure}

After constructing the MMTab dataset, we endeavor to fully leverage this data to promote models' multimodal table understanding ability. Inspired by the widely adopted training paradigm of previous MLLMs~\citep{blip2,liu2023llava_1,minigpt4}, we devise an enhanced two-stage training procedure and choose LLaVA-1.5~\citep{improved_llava} as the backbone to develop a versatile tabular MLLM named Table-LLaVA. The whole training process is illustrated in the Figure \ref{model_overview}.

\subsection{Model Architecture} Following LLaVA-1.5, the proposed Table-LLaVA consists of three modules: a pre-trained ViT model~\citep{clip_paper} as the visual encoder, a two-layer MLP as the vision-language connector and a Vicuna model~\citep{vicuna2023} as the backbone LLM. The ViT model encodes the input image into visual features, which are then projected into the word embedding space of LLM by the MLP connector. The Vicuna takes as input the concatenation of processed visual features and embedded textual features to generate responses. 

\subsection{Model Training} 
\paragraph{Pre-training.} As depicted in the top-left region of Fig. \ref{model_overview}, the vision-language connector is first pre-trained with an extra table recognition task on the MMTab-pre dataset, where the model is required to output a textual table representation (e.g., an HTML string) which encompasses both the table structure and table content. This process aims at aligning the visual features of diversified table images with the ground-truth textual table representations, which endows the model with augmented table structure perceiving and OCR ability and thus lays the foundation of more advanced tabular tasks.

\paragraph{Instruction fine-tuning.} In the second stage, the pre-trained vision-language connector and the LLM are jointly fine-tuned with instruction following data of multimodal table tasks in MMTab-instruct and traditional multimodal tasks. While a plethora of multimodal instruction following datasets have been previously constructed~\citep{liu2023llava_1,lyu2023macaw_llm,xu2023multi_instruct}, none of them have adequately solved the multimodal table understanding problem. The proposed MMTab-instruct contributes to addressing this gap and we use it to endow the model with the advanced ability to perform downstream table tasks. We also include the original pre-training and fine-tuning data of LLaVA-1.5 during the training process to improve the generalization of the resulting model and we analyze their influence in the ablation study.

\begin{table*}[t]\footnotesize
\centering
\renewcommand{\arraystretch}{1.3}
\setlength\tabcolsep{2pt}
\scalebox{0.78}{
\begin{tabular}{cccccccccccccc} 
\hline
\multirow{3}{*}{\textbf{Method}} & \multirow{3}{*}{\textbf{LLM}} & \multicolumn{1}{c|}{\multirow{3}{*}{\textbf{Res.}}} & \multicolumn{5}{c|}{\textbf{Question Answering}} & \multicolumn{2}{c|}{\textbf{Fact Verification}} & \multicolumn{4}{c}{\textbf{Text Generation}} \\ 
\cline{4-14}
 &  & \multicolumn{1}{c|}{} & \multicolumn{1}{c|}{\textbf{TABMWP}} & \multicolumn{1}{c|}{\textbf{WTQ}} & \multicolumn{1}{c|}{\textbf{HiTab}} & \multicolumn{1}{c|}{\textbf{TAT-QA}} & \multicolumn{1}{c|}{\textbf{FeTaQA}} & \multicolumn{1}{c|}{\textbf{TabFact}} & \multicolumn{1}{c|}{\textbf{InfoTabs}} & \multicolumn{1}{c|}{\textbf{ToTTo}} & \multicolumn{1}{c|}{\textbf{HiTab\_T2T}} & \multicolumn{1}{c|}{\textbf{Rotowire}} & \textbf{WikiBIO} \\ 
\cline{4-14}
 &  & \multicolumn{1}{c|}{} & \multicolumn{1}{c|}{Acc.} & \multicolumn{1}{c|}{Acc.} & \multicolumn{1}{c|}{Acc.} & \multicolumn{1}{c|}{Acc.} & \multicolumn{1}{c|}{BLEU} & \multicolumn{1}{c|}{Acc.} & \multicolumn{1}{c|}{Acc.} & \multicolumn{1}{c|}{BLEU} & \multicolumn{1}{c|}{BLEU} & \multicolumn{1}{c|}{BLEU} & BLEU \\ 
\hline
\multicolumn{14}{l}{{\cellcolor[rgb]{0.957,0.957,0.957}}\textit{MLLM}} \\
BLIP & 385M & 384 & 3.94 & 1.24 & 0.12 & 0.13 & 0.02 & 0.17 & 0.22 & 0 & 0.18 & 0.04 & 0.02 \\
OFA-Huge & 930M & - & 0 & 0.06 & 0.07 & 0 & 0.07 & 0.26 & 0.11 & 0.20 & 0.15 & 0 & 0 \\
BLIP2 & Flan-T5 3B & 224 & 3.34 & 2.01 & 1.52 & 2.20 & 2.34 & 18.62 & 27.53 & 4.3 & 2.63 & 1.08 & 0.72 \\
MiniGPT-4 & Vicuna 7B & 224 & 0.22 & 0.90 & 0.20 & 0.13 & 0.39 & 0 & 0.10 & 0.20 & 0.11 & 1.26 & 0.33 \\
Qwen-VL & Qwen 7B & 448 & 3.30 & 0.09 & 0.06 & 0.13 & 0.45 & 1.12 & 0.65 & 0.80 & 0.18 & 0 & 0 \\
InternLM-XComposer & InternLM 7B & 224 & 0.06 & 0.05 & 0.12 & 0.26 & 2.62 & 1.19 & 1.11 & 7.10 & 3.25 & 0.43 & 1.52 \\
mPLUG-Owl & Llama 7B & 224 & 1.76 & 0.62 & 0.25 & 0.13 & 7.42 & 7.46 & 5.53 & 3.50 & 1.75 & 1.96 & 1.37 \\
mPLUG-Owl2 & Llama-2 7B & 448 & 6.83 & 0.67 & 0.13 & 0.39 & 11.91 & 8.21 & 26.19 & 5.30 & 2.11 & 1.23 & 2.16 \\
LLaVA v1.5 & Vicuna-1.5 7B & 336 & 6.05 & 1.24 & 2.03 & 2.97 & 8.24 & 18.9 & 28.31 & 6.40 & 2.07 & 1.92 & 2.34 \\
Vary-toy & Qwen 1.8B & 1024 & 4.42 & 7.96 & 3.42 & 8.81 & 2.44 & 6.33 & 6.98 & 0.70 & 0.27 & 0.46 & 0.37 \\
Monkey & Qwen 7B & 896 & 13.26 & 19.07$^\dagger$ & 6.41 & 12.31 & 3.41 & 22.56$^\dagger$ & 22.11 & 3.50 & 1.12 & 0.03 & 2.77 \\
\multicolumn{14}{l}{{\cellcolor[rgb]{0.957,0.957,0.957}}\textit{LLM}} \\
\textit{Llama2+Oracle} & \textit{Llama-2 7B} & \textit{-} & \textit{17.88} & \textit{4.26} & \textit{1.21} & \textit{3.62} & \textit{5.54} & \textit{4.21} & \textit{7.55} & \textit{6.20} & \textit{1.84} & \textit{4.67} & \textit{1.33} \\
Llama2+OCR & Llama-2 7B & - & 16.35 & 3.91 & 0.77 & 5.27 & 5.15 & 4.32 & 7.17 & - & 1.56 & 3.90 & 1.28 \\
\textit{TableLlama+Oracle} & \textit{Llama-2 7B} & \textit{-} & \textit{12.98} & \textit{31.63}$^\ddagger$ & \textit{64.71}$^\ddagger$ & \textit{2.84} & \textit{39.05}$^\ddagger$ & \textit{82.55}$^\ddagger$ & \textit{2.85} & \textit{20.77}$^\ddagger$ & \textit{0.19} & \textit{0.13} & \textit{0.39} \\
TableLlama+OCR & Llama-2 7B & - & 11.09 & 12.49 & \textbf{13.51}$^\dagger$ & 2.72 & 25.44$^\dagger$ & 44.54$^\dagger$ & 2.18 & - & 0.12 & 0.13 & 0.31 \\
\multicolumn{14}{l}{{\cellcolor[rgb]{0.957,0.957,0.957}}\textit{Ours}} \\
\textbf{Table-LLaVA} 7B & Vicuna-1.5 7B & 336 & 57.78 & 18.43 & 10.09 & 12.82 & 25.60 & 59.85 & 65.26 & 23.00 & 9.74 & \textbf{10.46} & \textbf{9.68} \\
\textbf{Table-LLaVA} 13B & Vicuna-1.5 13B & 336 & \textbf{59.77} & \textbf{20.41} & 10.85 & \textbf{15.67} & \textbf{28.03} & \textbf{65.00} & \textbf{66.91} & \textbf{24.10} & \textbf{10.40} & 8.83 & 9.67 \\
\hline
\end{tabular}
}
\caption{Evaluation results on 11 held-in academic tabular benchmarks. `\textit{+Oracle}' and `+OCR' represents that the ground truth or OCR-extracted textual table representations are provided to LLMs, respectively. We only report model performance in the ideal `\textit{+Oracle}' setting and compare with models in the more practical `+OCR' setting. $\dagger$ indicates the model has been trained on the corresponding dataset, $\ddagger$ denotes results from original papers. } \label{OATB}
\end{table*}

\begin{table*}[t]\footnotesize
\centering
\renewcommand{\arraystretch}{1.2}
\setlength\tabcolsep{3pt}
\scalebox{0.95}{
\begin{tabular}{ccccccccccccc} 
\hline
\multirow{2}{*}{\textbf{Method}} & \multirow{2}{*}{\textbf{LLM}} & \multicolumn{1}{c|}{\multirow{2}{*}{\textbf{Res.}}} & \multicolumn{2}{c|}{\textbf{TSD}} & \multicolumn{1}{c|}{\textbf{TCE}} & \multicolumn{1}{c|}{\textbf{TCL}} & \multicolumn{1}{c|}{\textbf{MCD}} & \multicolumn{2}{c|}{\textbf{RCE}} & \multicolumn{3}{c}{\textbf{TR}} \\ 
\cline{4-13}
 &  & \multicolumn{1}{c|}{} & \multicolumn{1}{c|}{\begin{tabular}[c]{@{}c@{}}Row\\~Acc.\end{tabular}} & \multicolumn{1}{c|}{\begin{tabular}[c]{@{}c@{}}Col. \\Acc.\end{tabular}} & \multicolumn{1}{c|}{Acc.} & \multicolumn{1}{c|}{Acc.} & \multicolumn{1}{c|}{F1} & \multicolumn{1}{c|}{\begin{tabular}[c]{@{}c@{}}Row\\~F1\end{tabular}} & \multicolumn{1}{c|}{\begin{tabular}[c]{@{}c@{}}Col. \\F1\end{tabular}} & \multicolumn{1}{c|}{\begin{tabular}[c]{@{}c@{}}HTML\\TEDS\end{tabular}} & \multicolumn{1}{c|}{\begin{tabular}[c]{@{}c@{}}Markdown\\TEDS\end{tabular}} & \begin{tabular}[c]{@{}c@{}}Latex\\TEDS\end{tabular} \\ 
\hline
\multicolumn{13}{l}{{\cellcolor[rgb]{0.957,0.957,0.957}}\textit{MLLM}} \\
BLIP & 385M & 384 & 0 & 0.10 & 0.76 & 0 & 0 & 0 & 0 & 0 & 0.18 & 0 \\
OFA-Huge & 930M & - & 0 & 0.10 & 0.26 & 0 & 0 & 0 & 0 & 0 & 0.16 & 0 \\
BLIP2 & Flan-T5 3B & 224 & 0.20 & 0.30 & 0.15 & 0 & 0 & 0.06 & 0 & 0 & 0.25 & 0 \\
MiniGPT-4 & Vicuna 7B & 224 & 0.40 & 0.40 & 0 & 0 & 0 & 0 & 0 & 0 & 0.34 & 0 \\
Qwen-VL & Qwen 7B & 448 & 0 & 0 & 0.03 & 0.03 & 0.38 & 0 & 0 & 0 & 2.51 & 0 \\
InternLM-XComposer & InternLM 7B & 224 & 0.90 & 3.00 & 0.89 & 0.28 & 0.14 & 0.28 & 0.25 & 13.33 & 2.61 & 1.34 \\
mPLUG-Owl & Llama 7B & 224 & 1.20 & 3.90 & 0.13 & 0.16 & 0.34 & 2.04 & 1.38 & 15.31 & 7.36 & 3.13 \\
mPLUG-Owl2 & Llama-2 7B & 448 & 0.50 & 3.50 & 0.51 & 0.17 & 0.45 & 3.49 & 2.38 & 15.71 & 6.67 & 4.43 \\
LLaVA v1.5 & Vicuna-1.5 7B & 336 & 0.80 & 2.50 & 0.22 & 0.62 & 1.26 & 1.66 & 4.13 & 12.88 & 10.74 & 1.55 \\
Vary-toy & Qwen 1.8B & 1024 & 1.30 & 2.20 & 1.96 & 0.73 & 0.52 & 2.01 & 2.38 & 10.13 & 12.72 & 11.67 \\
Monkey & Qwen 7B & 896 & 0.80 & 0.60 & 1.46 & 1.31 & 0.67 & 3.89 & 4.53 & 21.96 & 13.29 & 4.54 \\
\multicolumn{13}{l}{{\cellcolor[rgb]{0.957,0.957,0.957}}\textit{LLM}} \\
\textit{Llama2+Oracle} & \textit{Llama-2 7B} & \textit{-} & \textit{1.70} & \textit{3.60} & \textit{0.62} & \textit{0.17} & \textit{-} & \textit{9.36} & \textit{18.03} & \textit{-} & \textit{-} & \textit{-} \\
Llama2+OCR & Llama-2 7B & - & 1.30 & 3.40 & 0.35 & 0.15 & - & 8.15 & 10.45 & - & - & - \\
\textit{TableLlama+Oracle} & \textit{Llama-2 7B} & \textit{-} & \textit{5.30} & \textit{4.40} & \textit{9.35} & \textit{0.82} & \textit{-} & \textit{4.34} & \textit{5.26} & \textit{-} & \textit{-} & \textit{-} \\
TableLlama+OCR & Llama-2 7B & - & 3.90 & 3.70 & 3.95 & 0.65 & - & 2.82 & 2.39 & - & - & - \\
\multicolumn{13}{l}{{\cellcolor[rgb]{0.957,0.957,0.957}}\textit{Ours}} \\
\textbf{Table-LLaVA 7B} & Vicuna-1.5 7B & 336 & 33.10 & \textbf{33.20} & 19.45 & 29.31 & \textbf{17.14} & \textbf{31.43} & 37.93 & 50.24 & 44.82 & 46.11 \\
\textbf{Table-LLaVA 13B} & Vicuna-1.5 13B & 336 & \textbf{34.40} & 27.60 & \textbf{19.53} & \textbf{29.68} & 16.52 & 31.07 & \textbf{41.49} & \textbf{51.44} & \textbf{46.00} & \textbf{46.50} \\
\hline
\end{tabular}
}
\caption{Evaluation results on 6 held-in table structure understanding benchmarks. For all evaluation metrics, higher values indicate better performance. HTML, Markdown and Latex represents the format of target textual table representations in the table recognition (TR) task, and TEDS is its evaluation metric. See Section \ref{experimental_setup} for the details. } \label{TSU}
\end{table*}

\begin{table}[t]\footnotesize
\centering
\renewcommand{\arraystretch}{1.2}
\setlength\tabcolsep{3pt}
\scalebox{0.87}{
\begin{tabular}{cccc|c|c} 
\hline
\textbf{Method} & \textbf{TQA} & \textbf{TFV} & \textbf{T2T} & \textbf{TSU} & \textbf{Held-out} \\ 
\hline
\multicolumn{6}{l}{{\cellcolor[rgb]{0.957,0.957,0.957}}GPT-4V (On a subset of test samples)} \\
Low Resolution & 24.15 & 52.00 & 2.42 & 28.11 & 30.40 \\
High Resolution & \textbf{35.91} & 55.55 & 3.05 & 31.16 & \textbf{44.49} \\
\multicolumn{6}{l}{{\cellcolor[rgb]{0.957,0.957,0.957}}Ours (On a subset of test samples)} \\
Table-LLaVA 7B & 24.55 & \textbf{65.25} & \textbf{9.49} & 34.24 & 23.16 \\
Table-LLaVA 13B & 26.63 & 64.50 & 9.12 & \textbf{34.36} & 24.71 \\
\hline \hline
\multicolumn{1}{l}{Table-LLaVA 13B} & 26.95 & 65.96 & 13.25 & 34.42 & 25.62 \\
\multicolumn{1}{l}{Table-LLaVA 7B} & 24.94 & \textbf{62.56} & \textbf{13.22} & \textbf{34.27} & \textbf{24.46} \\ 
\hline
\multicolumn{1}{l}{w/o LLaVA-pre} & 24.06 & 61.45 & 12.40 & 31.18 & 21.50 \\
$\triangle$ & {\cellcolor[rgb]{1,0.914,0.914}}-0.88 & {\cellcolor[rgb]{1,0.812,0.812}}-1.11 & {\cellcolor[rgb]{1,0.914,0.914}}-0.82 & {\cellcolor[rgb]{0.965,0.753,0.753}}-3.09 & {\cellcolor[rgb]{0.965,0.753,0.753}}-2.96 \\ 
\cline{2-6}
\multicolumn{1}{l}{w/o MMTab-pre} & 23.45 & 60.32 & 12.26 & 29.55 & 21.73 \\
$\triangle$ & {\cellcolor[rgb]{1,0.812,0.812}}-1.49 & {\cellcolor[rgb]{0.965,0.753,0.753}}-2.24 & {\cellcolor[rgb]{1,0.914,0.914}}-0.97 & {\cellcolor[rgb]{1,0.592,0.592}}-4.73 & {\cellcolor[rgb]{0.965,0.753,0.753}}-2.72 \\ 
\hline
\multicolumn{1}{l}{w/o LLaVA-instruct} & \textbf{24.98} & 61.85 & 12.87 & 33.98 & 23.90 \\
$\triangle$ & {\cellcolor[rgb]{0.894,1,0.894}}+0.04 & {\cellcolor[rgb]{1,0.914,0.914}}-0.71 & {\cellcolor[rgb]{1,0.914,0.914}}-0.36 & {\cellcolor[rgb]{1,0.914,0.914}}-0.29 & {\cellcolor[rgb]{1,0.914,0.914}}-0.56 \\ 
\cline{2-6}
\multicolumn{1}{l}{w/o MMTab-instruct} & 2.82 & 20.57 & 4.08 & 5.68 & 3.02 \\
$\triangle$ & {\cellcolor[rgb]{0.902,0.506,0.506}}-22.12 & {\cellcolor[rgb]{1,0.4,0.4}}-41.99 & {\cellcolor[rgb]{0.949,0.533,0.533}}-9.14 & {\cellcolor[rgb]{1,0.537,0.537}}-28.60 & {\cellcolor[rgb]{0.902,0.506,0.506}}-21.43 \\ 
\cline{2-6}
w/o TSU-instruct & 24.34 & 62.28 & 12.39 & 5.99 & 13.24 \\
$\triangle$ & {\cellcolor[rgb]{1,0.914,0.914}}-0.60 & {\cellcolor[rgb]{1,0.914,0.914}}-0.28 & {\cellcolor[rgb]{1,0.914,0.914}}-0.83 & {\cellcolor[rgb]{1,0.537,0.537}}-28.28 & {\cellcolor[rgb]{1,0.592,0.592}}-11.22 \\ 
\hline
\multicolumn{1}{l}{w successively IFT} & 24.76 & 61.99 & 13.06 & 33.89 & 23.85 \\
$\triangle$ & {\cellcolor[rgb]{1,0.914,0.914}}-0.18 & {\cellcolor[rgb]{1,0.914,0.914}}-0.57 & {\cellcolor[rgb]{1,0.914,0.914}}-0.16 & {\cellcolor[rgb]{1,0.914,0.914}}-0.38 & {\cellcolor[rgb]{1,0.914,0.914}}-0.61 \\ 
\hline 
\end{tabular}
}

\caption{Upper: Comparison with GPT-4V. Lower: Ablation experiment results. We report average performance over benchmarks under five types, respectively. $\triangle$ stands for the performance gap between Table-LLaVA 7B and its variants. `TSU-instruct' stands for 6 table structure understanding datasets in MMTab-instruct. `successively IFT' represents that `LLaVA-instruct' and `MMTab-instruct' are used to fine-tune the model in a sequential order rather than mixed together. }\label{ablation}

\end{table}

\section{Experiments}
\label{experiments}

\subsection{Experimental Setup}
\label{experimental_setup}

\textbf{Baselines.} We consider baselines of three genres: (1) Open-source MLLMs including BLIP~\citep{li2022blip_model}, OFA-Huge~\citep{wang2022ofa_model}, BLIP2~\citep{blip2}, MiniGPT-4~\citep{minigpt4}, Qwen-VL~\citep{bai2023qwenvl_model}, InternLM-XComposer~\citep{zhang2023internlmxcomposer}, mPLUG-Owl~\citep{ye2023mplugowl} and mPLUG-Owl2~\citep{ye2023mplugowl2}, LLaVA-1.5~\citep{improved_llava}, Vary-toy~\citep{wei2024vary_toy} and Monkey~\citep{li2023monkey}. (2) Open-source LLMs including Llama2~\citep{touvron2023llama2_model} and its counterpart TableLlama~\citep{tablellama}, which uses LongLoRA~\citep{chen2023longlora} to instruction-tune LLama2 on a series of textual tabular tasks. (3) The GPT-4V with low and high image resolution. Considering the high cost of GPT-4V, we randomly select 100 or 200 samples from each benchmark to compare Table-LLaVA with GPT-4V. To enable LLMs to digest table images, we consider an ideal scenario where LLMs are provided with oracle table sequences to explore the performance upper bound, and a more practical scenario where available table sequences are recognized from images by a competitive OCR engine~\citep{paddleocr_table_recognition}. For all methods, the zero-shot setting was adopted during evaluation. Implementation details can be found in App. \ref{more_implementation_details}.

\textbf{Evaluation metrics.} For TQA, TFV, and T2T benchmarks, we use accuracy or BLEU~\citep{bleu_metric}. For TSD, we compute accuracy for predicted row and column numbers separately. For TCE and TCL, we compute accuracy at cell-level. For MCD, we use cell-level F1. For RCE, we compute cell-level F1 for extracted rows and columns, respectively. For table recognition (TR) task, we follow \citet{pubtabnet} and use the Tree-Edit-Distance-based Similarity (TEDS) score, which is based on the tree structure of HTML table sequence and can measure both the structure similarity and the cell content similarity between the prediction and the ground truth. The score is normalized between 0 and 1, where 1 means perfect matching. For TR testing samples whose target sequences are in the Markdown or Latex format, we convert the predicted sequences into the HTML format to compute their TEDS scores.

\subsection{Results and Analysis}
\label{results_and_analysis}

\paragraph{Public academic tabular benchmark results.}
\textit{Performance of open-source MLLMs.}
As we can see from the MLLM rows in Table \ref{OATB}, early MLLMs (e.g., MiniGPT-4, BLIP)  exhibited minimal proficiency in multimodal table understanding due to the lack of tabular training data, but recent MLLMs (e.g., LLaVA-1.5 and Monkey) have yielded better capacity for comprehending table images, which can be attributed to their improvements on the OCR and text-rich scenarios. Especially, among existing MLLMs, Monkey performs the best in most question answering tasks and fact verification tasks because of the training on relevant table datasets (i.e., WTQ and TabFact).

\textit{Performance of LLMs.} As shown in Table \ref{OATB}, TableLlama+OCR performs better than Llama2+OCR on several tasks (e.g., HiTab, FeTaQA, TabFact) through fine-tuning on the corresponding training data, but this also damages its generalization ability on unseen tasks (e.g., InfoTabs and TABMWP). Compared to Llama2+OCR, Llama2+Oracle does not achieve notable improvements, indicating that its bottleneck is the ability to understand tables and follow related instructions, rather than the table recognition ability. On the contrary, TableLlama+Oracle consistently outperforms TableLlama+OCR in all tasks, because it has been instruction-tuned on large-scale tabular data, which leads to better table understanding and instruction-following ability. Thus, the provided oracle table sequences break the bottleneck of the OCR engine's table recognition capability, resulting in significant improvements.

\textit{Comparison between Table-LLaVA and existing  models.} Compared to previous open-source MLLMs and LLMs+OCR, Table-LLaVA 7B and 13B both surpass them with large margins, demonstrating the effectiveness of our methods and the value of MMTab dataset. One exception is the accuracy of TableLlama+OCR on HiTab, which maybe because table images in this benchmark are relatively large, leading to information loss when resizing them into desired resolutions of Table-LLaVA (i.e., 336$\times$336). We believe there is great potential for using more powerful MLLMs to perform diverse multimodal table understanding tasks.

\paragraph{Table structure understanding benchmark results.}

Table structure understanding is a fundamental ability for fulfilling more advanced tabular tasks. As can been found in Table \ref{TSU}, both previous MLLMs and LLMs failed to generalize well on these relatively simple tabular benchmarks that are almost trivial for humans. What's more, their performance is even worse than that on more challenging academic benchmarks in Table \ref{OATB}. This shows that these powerful (M)LLMs may rely on some superficial correlations~\citep{shortcut_learning_in_dnn} to perform downstream tabular tasks that require complex reasoning, and they actually lack the important ability to perceive basic table structures.

\paragraph{Held-out tabular benchmark results.}
Table \ref{heldout_benchmark} reports model performance on 7 held-out benchmarks whose data do not appear in the model training. We can find that previous open-source models excel at different benchmarks respectively, and no model can consistently outperform others on all these tasks. By contrast, Table-LLaVA achieves best performance on most benchmarks, except for the accuracy of Vary-toy on AIT-QA, which is because AIT-QA contains large tables extracted from annual reports of airline companies and Vary-toy might have seen similar tables in its training data of company document images. Besides, Vary-toy supports higher input image resolution (1024), which is more friendly for large tables.

\paragraph{Comparison with GPT-4V.} The average performance of Table-LLaVA and GPT-4V on five types of benchmarks is shown in the upper part of Table \ref{ablation}. GPT-4V achieves remarkable results under both low (512$\times$512) and high (768$\times$2000) image resolution. The average performance of Table-LLaVA 7B (336$\times$336) is better than GPT-4V with low resolution (512$\times$512) on four types of benchmarks, while GPT-4V surpasses Table-LLaVA in the held-out scenario, indicating its strong generalization ability. As can be seen from detailed benchmark performance in Table \ref{OATB_GPT4V}, Table \ref{TSU_GPT4V} and Table \ref{heldout_benchmark}, Table-LLaVA achieves better or competitive results with GPT-4V on 14 out of 24 benchmarks. Besides, GPT-4V can obtain significant improvements from high image resolution, which helps the model comprehend fine-grained table elements and structures in large tables. We also analyze the influence of input image resolution on the performance of Table-LLaVA in Appendix \ref{influence_of_image_resolution}.

\paragraph{Ablation study.}
We conduct sufficient ablation experiments to validate the effectiveness of our proposed dataset and training strategy. We divide the ablation study into three parts: (1) \textit{Ablation of pre-training.} As shown in Table \ref{ablation}, both `w/o LLaVA-pre' and  `w/o MMTab-pre'  cause negative effects, and the latter results in a larger decline. This is because both LLaVA-pre and MMTab-pre help align visual and textual modalities, while MMTab-pre is more suitable for multimodal alignment of table understanding. (2) \textit{Ablation of instruction fine-tuning.} `w/o LLaVA-instruct' causes a slight performance decrease, indicating that though LLaVA-instruct has different image domains and task settings from MMTab, it has benefits for the multimodal table understanding due to the enhancement of instruction-following ability. `w/o MMTab-instruct' leads to a significant performance drop on all types of tasks, resulting in extremely poor performance (e.g., 3.02 on held-out benchmarks). This further confirms that our constructed data can supplement the missing table understanding capability of the current MLLMs. If the table structure understanding data in MMTab-instruct is removed (i.e., `w/o TSU-instruct'), we can find that, although it does not cause obvious performance damage to traditional academic tasks like TQA and TFV, it has a huge negative impact on TSU and Held-out tasks. This indicates that the proposed TSU datasets also help with model generalization. (3) \textit{Ablation of  training strategies.} We compare models instruction-tuned with LLaVA-pre and MMTab-pre in sequence (`w successfully IFT') or mixed together. We find that `w successfully IFT' has slightly weaker performance, which suggests that mixed data is more conducive to model performance.

\paragraph{The influence of MMTab on non-tabular tasks.} Table \ref{results_of_table_llava_on_non_tabular_tasks} lists performance of Table-LLaVA and its backbone LLaVA-1.5 on two non-tabular benchmarks: TextVQA~\citep{textvqa_benchmark} and LLaVA-Bench (In-the-Wild)~\citep{liu2023llava_1}. Table-LLaVA beats LLaVA-1.5 in most cases under both model sizes, which demonstrates that MMTab actually has positive impact on the performance of non-tabular tasks. Combing this with ablation of non-tabular training data, we can find that there are mutual benefits between model's capacity for tabular tasks and non-tabular tasks, which shows that table understanding is one fundamental and necessary ability of MLLM and it deserves more investigations. \textbf{More results and analysis such as case study are shown in Appendix \ref{more_experimental_results}}.

\section{Conclusion}

This paper proposes a novel multimodal table understanding problem, together with a large-scale open-source dataset MMTab, which covers a broad range of table structures and tabular tasks. This dataset provides a comprehensive testbed for MLLM research with held-in and held-out multimodal tabular benchmarks. On the basis of MMTab, we empower LLaVA 1.5 to be a generalist tabular MLLM Table-LLaVA. Experimental results show that Table-LLaVA significantly outperforms existing MLLMs on multiple benchmarks, and is even on par with the powerful GPT-4V. In conclusion, the contributions of this paper lie at promoting the research on multimodal table understanding from the task, dataset and model perspectives.

\section{Limitations}
Though this work makes the first comprehensive exploration towards the multimodal table understanding problem, there are certain limitations that can be left to the follow-ups. First, the proposed dataset mainly focus on the single table in English. The multi-table scenario together with broader language coverage have not yet been considered. Second, MMTab is based on real-world tables from carefully selected table datasets and it contains diverse high-quality table images rendered by automatic scripts. Nevertheless, table images in the wild can be low-quality. For instance, blurred, handwritten or incomplete table images. To further bridge the gap between the academic research and the real application scenarios, more diversified table images from the wild could be collected in the future, and their corresponding instruction following data needs to be constructed. We believe this could significantly promote the applications of MLLM-based table understanding systems. In the end, though the proposed Table-LLaVA demonstrates great performance on a wide range of table-based tasks, the resolution of input images is relatively low and may limit the upper bound of its capacity. Luckily, with the emergence of MLLMs which possess higher input image resolution (e.g., Monkey~\citep{li2023monkey}, LLaVA-Next~\citep{liu2024llava_next}), we can use MMTab to develop more powerful tabular MLLM in the future research.

\section{Ethical Considerations}
\label{ethical considerations}
The proposed MMTab dataset is constructed based on the academic datasets like WTQ and TabFact, which are free and open datasets for research use with licenses like MIT License\footnote{https://opensource.org/license/mit/} or CC-BY-SA-4.0 License \footnote{https://creativecommons.org/licenses/by-sa/4.0/deed.en}. We write Python scripts to render textual table sequences (like HTML) in these datasets to obtain table images, and build multimodal instruction-following data based on original samples. The resulting dataset MMTab is also a free and open resource for the community to study the multimodal table understanding problem. Thus, the authors foresee no ethical concerns with the research in this paper.


\bibliography{custom}

\newpage
\appendix

\section{More Details about MMTab}
\label{more_details_of_dataset_construction}
\subsection{Task Descriptions and More Dataset Examples}
\label{more_dataset_examples}

\begin{figure*}[t]
  \centering
  \includegraphics[width=0.87\linewidth]{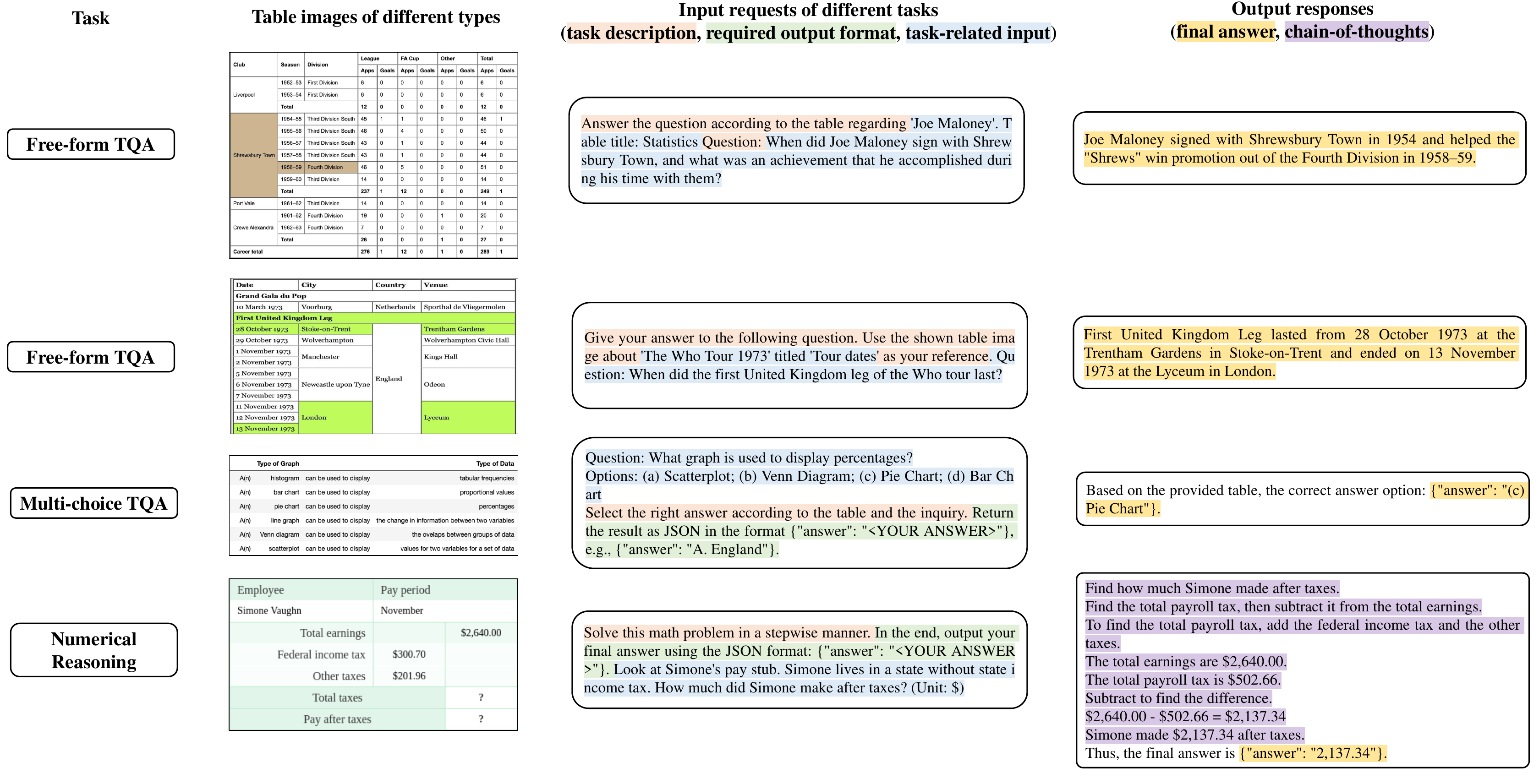}
  \caption{More dataset examples. 
  }
  \label{more_dataset_example_1}
  
\end{figure*}

\begin{figure*}[t]
  \centering
  \includegraphics[width=\linewidth]{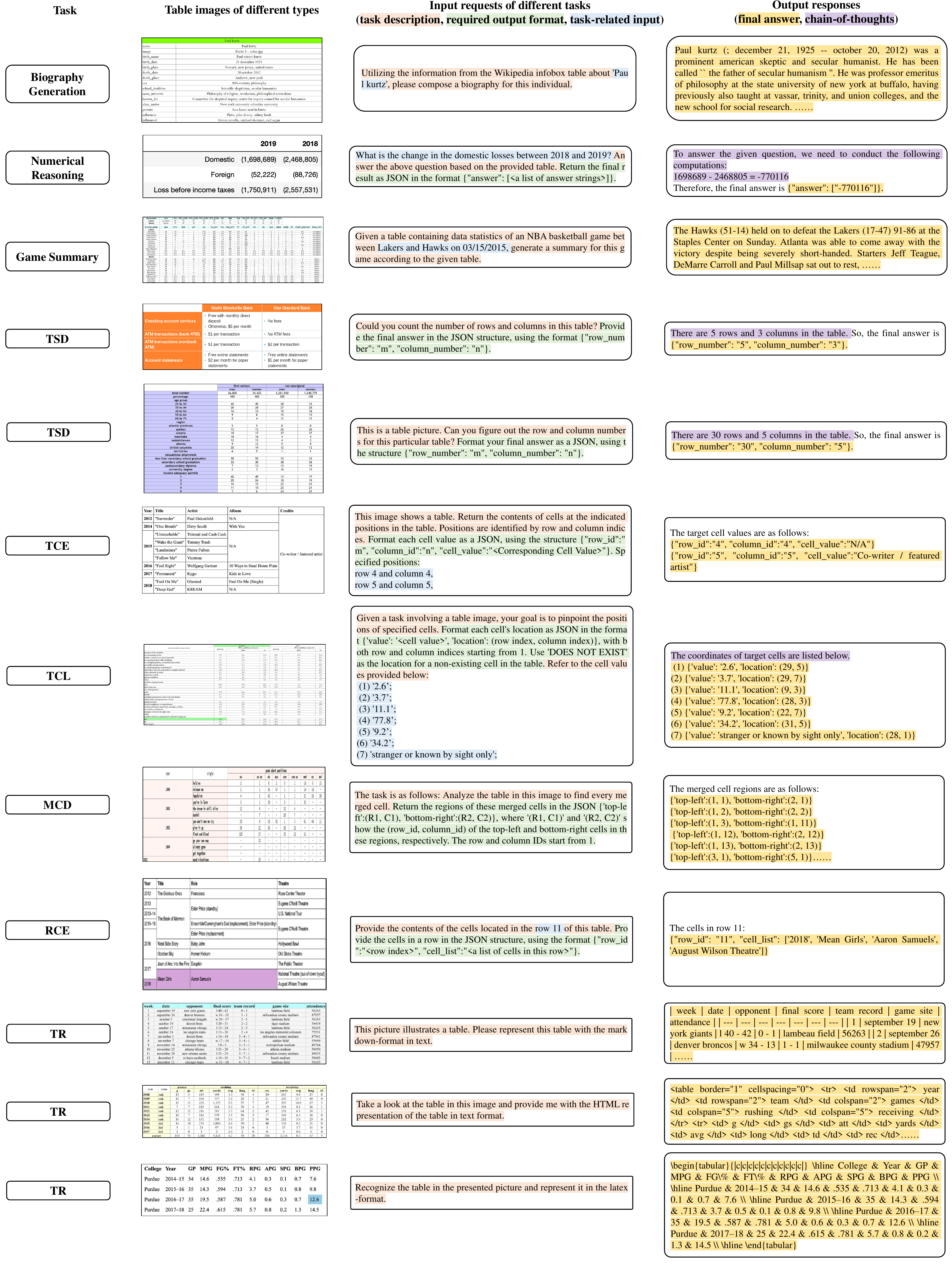}
  \caption{More dataset examples. 
  }
  \label{more_dataset_example_5}
  
\end{figure*}

\begin{figure*}[t]
  \centering
  \includegraphics[width=\linewidth]{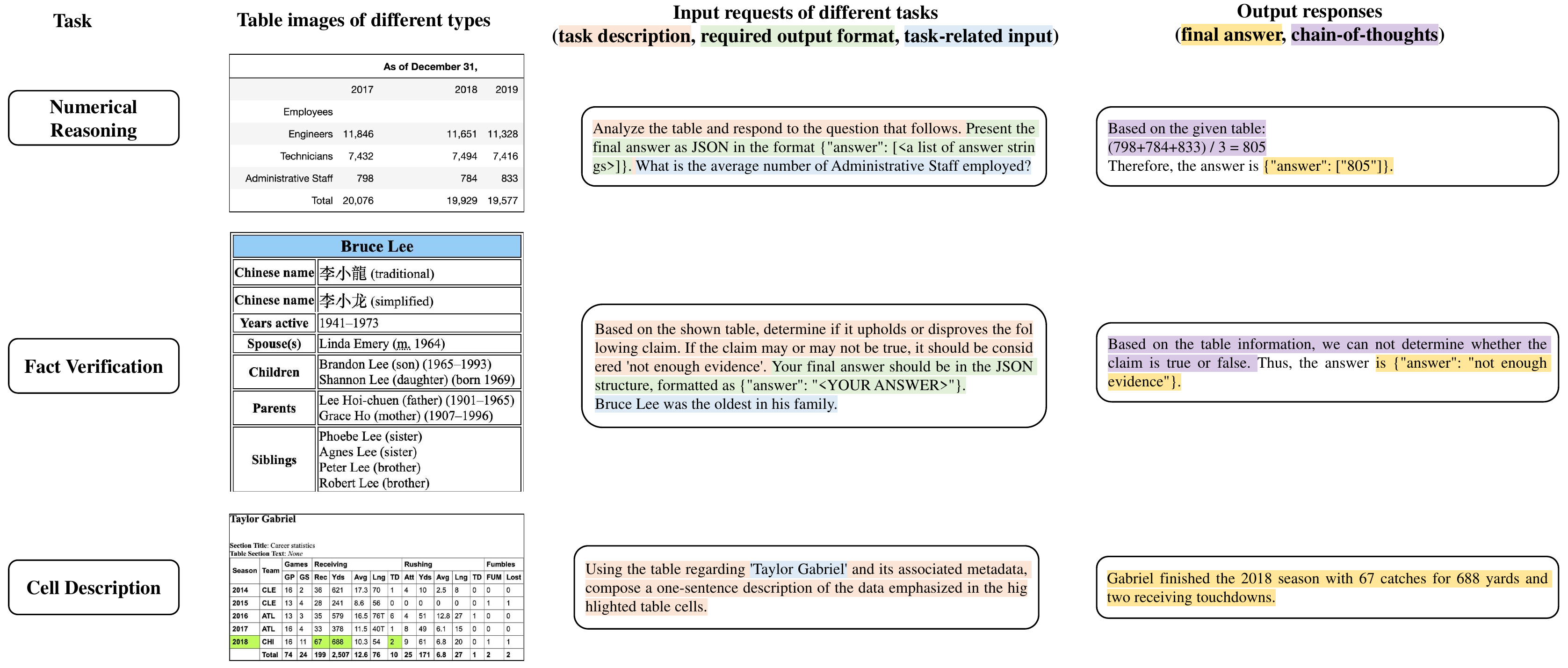}
  \caption{More dataset examples. 
  }
  \label{more_dataset_example_2}
  
\end{figure*}

Detailed description and evaluation metric of each task are given in Table \ref{task_definitions}, and more dataset examples are illustrated in Figure \ref{more_dataset_example_1}, \ref{more_dataset_example_5}, \ref{more_dataset_example_2}. When we collect tables from TabMCQ dataset, we filter extremely long tables more than 50 rows. For the hybrid-QA dataset TAT-QA, we only preserve samples whose questions can be answered with the table information. For the ToTTo dataset, its training set contains 35K tables and we randomly select 15K tables for MMTab-instruct in order to reduce the cost of transforming HTML tables into images. 

Except augmentation strategies mentioned in Section \ref{data_augmentations}, we also perform additional data augmentations including: (1) ``response-level augmentations'', where we synthesize chain-of-thoughts using annotated intermediate computational procedures in the original datasets and use them to augment the final answer. (2) ``conversation-level augmentations'', where we randomly choose samples of the same table to compose multi-turn conversation samples. 

\begin{table}[h]\footnotesize 
\centering
\renewcommand{\arraystretch}{1.3}
\setlength\tabcolsep{3pt}
\scalebox{0.75}{
\begin{tabular}{c|cc} 
\hline
Hyperparameter & Pre-train & Fine-tune \\ 
\hline
training data & \begin{tabular}[c]{@{}c@{}}MMTab-pre (150K),\\LLaVA-pre (558K)\end{tabular} & \begin{tabular}[c]{@{}c@{}}MMTab-instruct (232K),\\LLaVA-instruct (665K)\end{tabular} \\
batch size & 256 & 128 \\
max length & \multicolumn{2}{c}{2560} \\
learning rate (lr) & 1e-3 & 2e-5 \\
lr schedule & \multicolumn{2}{c}{cosine decay} \\
warmup ratio & \multicolumn{2}{c}{0.03} \\
weight decay & \multicolumn{2}{c}{0} \\
optimizer & \multicolumn{2}{c}{AdamW} \\
epoch & \multicolumn{2}{c}{1} \\
Deepspeed Stage & 2 & 3 \\
machine & \multicolumn{2}{c}{one machine with 8 80GB A800} \\
\begin{tabular}[c]{@{}c@{}}training time\\(w/o flash-attention)\end{tabular} & 32 hours & 26 hours \\
\hline
\end{tabular}
}
\caption{Hyperparameter setting and training details of Table-LLaVA.}
\label{hyperparameters}
\end{table}

\begin{table*}[t]\footnotesize
\centering
\renewcommand{\arraystretch}{1.3}
\setlength\tabcolsep{3pt}
\scalebox{0.70}{
\begin{tabular}{c|c|c|c|l|c} 
\hline
\textbf{MMTab} & \textbf{Task Category} & \textbf{Task Name} & \textbf{Dataset} & \multicolumn{1}{c|}{\textbf{Task Description}} & \textbf{Metric} \\ 
\hline
\multirow{20}{*}{\begin{tabular}[c]{@{}c@{}}MMTab-\\instruct\end{tabular}} & \multirow{7}{*}{\begin{tabular}[c]{@{}c@{}}Question \\Answering\end{tabular}} & \begin{tabular}[c]{@{}c@{}}Flat TQA\\~(F TQA)\end{tabular} & WTQ & \begin{tabular}[c]{@{}l@{}}TQA based on tables which usually possesses a flat \\structure with the first row as the sole column header.~~\end{tabular} & Accuracy($\uparrow$) \\ 
\cline{3-6}
 &  & Free-form TQA & FeTaQA & \begin{tabular}[c]{@{}l@{}}TQA with a free-form text answer rather than a\\~short text span copied from the table.\end{tabular} & BLEU($\uparrow$) \\ 
\cline{3-6}
 &  & \multirow{2}{*}{\begin{tabular}[c]{@{}c@{}}Hierarchical TQA\\(H TQA)\end{tabular}} & HiTab & \multirow{2}{*}{\begin{tabular}[c]{@{}l@{}}TQA based on tables which usually possesses\\~hierachical headers and merged cells.\end{tabular}} & Accuracy($\uparrow$) \\ 
\cline{4-4}\cline{6-6}
 &  &  & AIT-QA &  & Accuracy($\uparrow$) \\ 
\cline{3-6}
 &  & Multi-choice TQA & TabMCQ & TQA with multi-choice questions. & Accuracy($\uparrow$) \\ 
\cline{3-6}
 &  & \multirow{2}{*}{\begin{tabular}[c]{@{}c@{}}Tabular\\Numerical Reasoning\end{tabular}} & TABMWP & \multirow{2}{*}{\begin{tabular}[c]{@{}l@{}}TQA requiring mathematical reasoning operations such as \\finding the largest number or do math computations.\end{tabular}} & Accuracy($\uparrow$) \\ 
\cline{4-4}\cline{6-6}
 &  &  & TAT-QA &  & Accuracy($\uparrow$) \\ 
\cline{2-6}
 & \multirow{3}{*}{\begin{tabular}[c]{@{}c@{}}Fact \\Verification\end{tabular}} & \multirow{3}{*}{\begin{tabular}[c]{@{}c@{}}Table \\Fact~Verification\end{tabular}} & TabFact & \multirow{3}{*}{\begin{tabular}[c]{@{}l@{}}Given a table as evidence and a statement, the\\~task is to distinguish whether the given\\~statement is entailed or refuted by~the table.\end{tabular}} & Accuracy($\uparrow$) \\ 
\cline{4-4}\cline{6-6}
 &  &  & InfoTabs &  & Accuracy($\uparrow$) \\ 
\cline{4-4}\cline{6-6}
 &  &  & PubHealthTab &  & Accuracy($\uparrow$) \\ 
\cline{2-6}
 & \multirow{4}{*}{\begin{tabular}[c]{@{}c@{}}Text \\Generation\end{tabular}} & \multirow{2}{*}{Cell Description} & ToTTo & \begin{tabular}[c]{@{}l@{}}Generate a one-sentence description for the\\~highlighted table cells.\end{tabular} & BLEU($\uparrow$) \\ 
\cline{4-6}
 &  &  & HiTab\_T2T & \begin{tabular}[c]{@{}l@{}}Generate a one-sentence description for the\\~highlighted table cells using the provided\\~operators such as SUM, DIVISION.\end{tabular} & BLEU($\uparrow$) \\ 
\cline{3-6}
 &  & Game Summary & Rotowire & \begin{tabular}[c]{@{}l@{}}Given a table recording~box- and line-scores\\~of an NBA~game, the task is to generate a\\~detail game summary which is sourced from~rotowire.com.\textcolor[rgb]{0.122,0.129,0.157}{}\end{tabular} & BLEU($\uparrow$) \\ 
\cline{3-6}
 &  & Biography Generation & WikiBIO & \begin{tabular}[c]{@{}l@{}}Given a table containing information of a\\~person, the task is to generate a biography\\~to introduce this person.~\end{tabular} & BLEU($\uparrow$) \\ 
\cline{2-6}
 & \multirow{6}{*}{\begin{tabular}[c]{@{}c@{}}Structure\\ Understanding\end{tabular}} & Table Size Detection & TSD & \begin{tabular}[c]{@{}l@{}}Determine the row number and column\\~number of the given table.\end{tabular} & \begin{tabular}[c]{@{}c@{}}Accuracy at row \\or column level($\uparrow$)\end{tabular} \\ 
\cline{3-6}
 &  & Table Cell Extraction & TCE & \begin{tabular}[c]{@{}l@{}}Given a group of (row\_id, column\_id), the task\\~is to extract the corresponding table cells.\end{tabular} & Accuracy($\uparrow$) \\ 
\cline{3-6}
 &  & Table Cell Locating & TCL & \begin{tabular}[c]{@{}l@{}}Given a group of cells, the task is to find\\~positions of these cells in the table and return \\their position in theformat of (row\_id,~column\_id).\end{tabular} & Accuracy($\uparrow$) \\ 
\cline{3-6}
 &  & Merged Cell Detection & MCD & \begin{tabular}[c]{@{}l@{}}Determine whether the table contains\\~merged cells and return postions of top-left \\and bottom-right cells in the merged regions.\end{tabular} & F1($\uparrow$) \\ 
\cline{3-6}
 &  & Row\&Column Extraction& RCE & \begin{tabular}[c]{@{}l@{}}Given a group of row\_id or column\_id, the~task is to extract the~\\corresponding table cells in the target rows or target columns.\end{tabular} & \begin{tabular}[c]{@{}c@{}}F1 at row \\or column level($\uparrow$)\end{tabular} \\ 
\cline{3-6}
 &  & Table Recognition & TR & \multirow{2}{*}{\begin{tabular}[c]{@{}l@{}}Given a table image, the task is to return a textual representation \\of the table in the format of HTML, Markdown or Latex Same\end{tabular}} & \multirow{2}{*}{TEDS($\uparrow$)} \\ 
\cline{1-4}
\begin{tabular}[c]{@{}c@{}}MMTab-\\pre\end{tabular} & \multicolumn{2}{c|}{Table Recognition} & TR for pre-training &  &  \\
\hline
\end{tabular}
}

\caption{Detailed description of each task and their evaluation metrics.}
\label{task_definitions}
\end{table*}

\begin{table}[t]\footnotesize
\centering
\renewcommand{\arraystretch}{1.2}
\setlength\tabcolsep{3pt}
\scalebox{0.69}{
\begin{tabular}{c|c|cccc} 
\hline
\multirow{2}{*}{\textbf{Models}} & \multirow{2}{*}{\textbf{TextVQA}} & \multicolumn{4}{c}{\textbf{LLaVA-Bench (in-the-wild)}} \\ 
\cline{3-6}
 &  & Conversation & \begin{tabular}[c]{@{}c@{}}Detail \\description\end{tabular} & \begin{tabular}[c]{@{}c@{}}Complex \\reasoning\end{tabular} & Overall \\ 
\hline
LLaVA v1.5 7B & 58.2* & 54.3 & 49.6 & 72.4 & 61.4 \\
Table-LLaVA 7B & \textbf{59.2} & \textbf{58.3} & \textbf{50.9} & \textbf{73.2} & \textbf{63.2} \\ 
\hline
LLaVA v1.5 13B & 61.3* & 72.0 & \textbf{53.8} & 72.0 & 67.5 \\
Table-LLaVA 13B & \textbf{61.9} & \textbf{72.0} & 53.7 & \textbf{77.1} & \textbf{69.6} \\
\hline
\end{tabular}
}
\caption{Comparison of Table-LLaVA and its backbone on non-tabular tasks. $*$ indicates results are from the original LLaVA-1.5 paper.}
\label{results_of_table_llava_on_non_tabular_tasks}
\end{table}

\subsection{Instruction Templates}
\label{instruction_templates}
\begin{figure*}[t]
  \centering
  \includegraphics[width=0.95\linewidth]{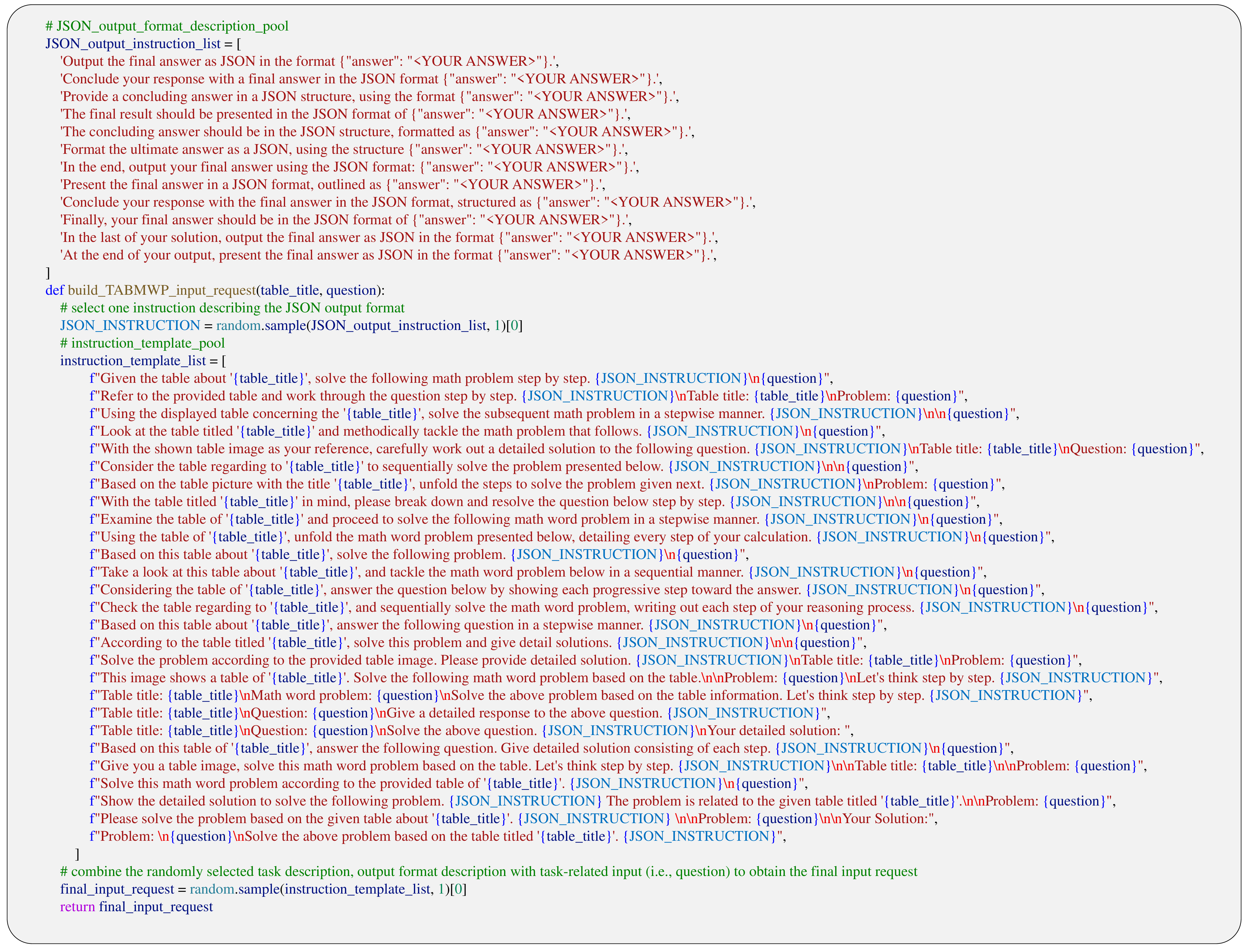}
  \caption{Exemplary instruction templates, JSON output format descriptions, and the Python Code for constructing the input requests. Taking the TABMWP dataset as an example. }
  \label{instruction_building_process}
  
\end{figure*}

The diversity of the instruction-following data has a significant impact on the performance of the resulting model. As discussed in the Section \ref{data_augmentations}, we utilize in-context learning to ask GPT-4 to generate new instruction templates and create more diversity of input request. When we build input requests of each dataset, we randomly choose an instruction template and an output format description from the candidate pool, and then combine them with the task-specific input such as the question to produce the final input request.  Figure \ref{instruction_building_process} shows the Python code for this combination process, together with all instruction templates and JSON output format descriptions for the TABMWP dataset. Previous textual instruction-following datasets for tabular tasks such as TableInstruct~\citep{tablellama} usually adopt one fixed instruction template for each dataset. By contrast, we construct at least 20 instruction templates for each dataset while considering their respective characteristics. 

\section{Implementation Details}
\label{more_implementation_details}
Following LLaVA-1.5~\citep{improved_llava}, we use the well-trained CLIP-ViT-L-336px~\citep{clip_paper} as the visual encoder and input images are resized to 336$\times$336. We develop two Table-LLaVA models with Vicuna-1.5 7B and 13B as the backbone LLM, and we denote the resulting models as Table-LLaVA 7B and Table-LLaVA 13B, respectively. We follow the original hyper-parameter setting of LLaVA-1.5 except that We increased the max sequence length from 2048 to 2560 to accommodate longer text sequences. The training hyperparameters for both the pre-training and the visual instruction tuning are listed in Table \ref{hyperparameters}. In this paper, all experiments including baseline experiments were conducted on a single machine with 8 80GB A800. Without using flash-attention~\citep{dao2022flashattention}, the pre-training process and the instruction-tuning takes about 32 hours and 26 hours for one epoch, respectively. Unless otherwise specified, we evaluate performance of baseline models on our benchmarks with the official implementations. As mentioned in the Section \ref{data_collection}, we add extra instructions to the input request which require models to output the final answer in the JSON format, and we write Python scripts with regular expressions to extract the final answer for a fair comparion. Some baselines like Monkey cannot follow instructions to output the answer in the desired JSON format,  which may only output a short answer due to the overfitting of their training data. Thus, we relaxed requirements and specifically designed answer extraction scripts to calculate their accuracy. For ToTTo benchmark, since the ground-truth of testing samples have not been open-sourced, we submit the output results of different models to the official website to get evaluation results. 

\section{More Experimental Results and Analysis}
\label{more_experimental_results}

\begin{table*}[t]\footnotesize
\centering
\renewcommand{\arraystretch}{1.3}
\setlength\tabcolsep{2pt}
\scalebox{0.75}{
\begin{tabular}{cccccccccccccc} 
\hline
\multirow{3}{*}{\textbf{Method}} & \multirow{3}{*}{\textbf{LLM}} & \multicolumn{1}{c|}{\multirow{3}{*}{\textbf{Res.}}} & \multicolumn{5}{c|}{\textbf{Question Answering}} & \multicolumn{2}{c|}{\textbf{Fact Verification}} & \multicolumn{4}{c}{\textbf{Text Generation}} \\ 
\cline{4-14}
 &  & \multicolumn{1}{c|}{} & \multicolumn{1}{c|}{\textbf{TABMWP}} & \multicolumn{1}{c|}{\textbf{WTQ}} & \multicolumn{1}{c|}{\textbf{HiTab}} & \multicolumn{1}{c|}{\textbf{TAT-QA}} & \multicolumn{1}{c|}{\textbf{FeTaQA}} & \multicolumn{1}{c|}{\textbf{TabFact}} & \multicolumn{1}{c|}{\textbf{InfoTabs}} & \multicolumn{1}{c|}{\textbf{ToTTo}} & \multicolumn{1}{c|}{\textbf{HiTab\_T2T}} & \multicolumn{1}{c|}{\textbf{Rotowire}} & \textbf{WikiBIO} \\ 
\cline{4-14}
 &  & \multicolumn{1}{c|}{} & \multicolumn{1}{c|}{Acc.} & \multicolumn{1}{c|}{Acc.} & \multicolumn{1}{c|}{Acc.} & \multicolumn{1}{c|}{Acc.} & \multicolumn{1}{c|}{BLEU} & \multicolumn{1}{c|}{Acc.} & \multicolumn{1}{c|}{Acc.} & \multicolumn{1}{c|}{BLEU} & \multicolumn{1}{c|}{BLEU} & \multicolumn{1}{c|}{BLEU} & BLEU \\ 
\hline
\multicolumn{14}{l}{{\cellcolor[rgb]{0.957,0.957,0.957}}\textit{Ours (on all test samples)}} \\
\textbf{Table-LLaVA} 7B & Vicuna-1.5 7B & 336 & 57.78 & 18.43 & 10.09 & 12.82 & 25.60 & 59.85 & 65.26 & 23.00 & 9.74 & \textbf{10.46} & \textbf{9.68} \\
\textbf{Table-LLaVA} 13B & Vicuna-1.5 13B & 336 & \textbf{59.77} & \textbf{20.41} & 10.85 & \textbf{15.67} & \textbf{28.03} & \textbf{65.00} & \textbf{66.91} & \textbf{24.10} & \textbf{10.40} & 8.83 & 9.67 \\ 
\hline
\multicolumn{14}{l}{{\cellcolor[rgb]{0.957,0.957,0.957}}\textit{GPT-4V (on a subset of~ test samples)}} \\
Low Resolution & GPT-4 & 512 & 60.00 & 22.50 & 9.50 & 19.50 & 9.26 & 45.50 & 58.50 & - & 1.85 & 3.89 & 1.55 \\
High Resolution & GPT-4 & 768*2000 & \textbf{60.50} & \textbf{48.00} & \textbf{27.50} & \textbf{32.50} & 11.04 & 45.50 & 65.60 & - & 2.98 & 4.23 & 1.94 \\
\multicolumn{14}{l}{{\cellcolor[rgb]{0.957,0.957,0.957}}\textit{Ours~(on a subset of~ test samples)}} \\
\textbf{Table-LLaVA~}7B & Vicuna-1.5 7B & 336 & 57.00 & 18.00 & 7.50 & 11.00 & 29.23 & \textbf{63.50} & 67.00 & - & 9.34 & \textbf{10.08} & \textbf{9.04} \\
\textbf{Table-LLaVA~}13B & Vicuna-1.5 13B & 336 & 60.00 & 21.50 & 8.00 & 14.00 & \textbf{29.63} & 59.50 & \textbf{69.50} & - & \textbf{9.53} & 9.00 & 8.84 \\
\hline
\end{tabular}
}
\caption{Comparison between GPT-4V and Table-LLaVA on 11 held-in public academic tabular benchmarks. Note that we randomly select a subset of testing samples for each tasks due to the high cost of GPT-4V and we also evaluate Table-LLaVA on the same subset. } \label{OATB_GPT4V}
\end{table*}

\begin{table*}[t]\footnotesize
\centering
\renewcommand{\arraystretch}{1.2}
\setlength\tabcolsep{3pt}
\scalebox{0.94}{
\begin{tabular}{ccccccccccccc} 
\hline
\multirow{2}{*}{\textbf{Method}} & \multirow{2}{*}{\textbf{LLM}} & \multicolumn{1}{c|}{\multirow{2}{*}{\textbf{Res.}}} & \multicolumn{2}{c|}{\textbf{TSD}} & \multicolumn{1}{c|}{\textbf{TCE}} & \multicolumn{1}{c|}{\textbf{TCL}} & \multicolumn{1}{c|}{\textbf{MCD}} & \multicolumn{2}{c|}{\textbf{RCE}} & \multicolumn{3}{c}{\textbf{TR}} \\ 
\cline{4-13}
 &  & \multicolumn{1}{c|}{} & \multicolumn{1}{c|}{\begin{tabular}[c]{@{}c@{}}Row\\~Acc.\end{tabular}} & \multicolumn{1}{c|}{\begin{tabular}[c]{@{}c@{}}Col. \\Acc.\end{tabular}} & \multicolumn{1}{c|}{Acc.} & \multicolumn{1}{c|}{Acc.} & \multicolumn{1}{c|}{F1} & \multicolumn{1}{c|}{\begin{tabular}[c]{@{}c@{}}Row\\~F1\end{tabular}} & \multicolumn{1}{c|}{\begin{tabular}[c]{@{}c@{}}Col. \\F1\end{tabular}} & \multicolumn{1}{c|}{\begin{tabular}[c]{@{}c@{}}HTML\\TEDS\end{tabular}} & \multicolumn{1}{c|}{\begin{tabular}[c]{@{}c@{}}Markdown\\TEDS\end{tabular}} & \begin{tabular}[c]{@{}c@{}}Latex\\TEDS\end{tabular} \\ 
\hline
\multicolumn{13}{l}{{\cellcolor[rgb]{0.957,0.957,0.957}}\textit{Ours~(on all test samples)}} \\
\textbf{Table-LLaVA 7B} & Vicuna-1.5 7B & 336 & 33.10 & \textbf{33.20} & 19.45 & 29.31 & \textbf{17.14} & \textbf{31.43} & 37.93 & 50.24 & 44.82 & 46.11 \\
\textbf{Table-LLaVA 13B} & Vicuna-1.5 13B & 336 & \textbf{34.40} & 27.60 & \textbf{19.53} & \textbf{29.68} & 16.52 & 31.07 & \textbf{41.49} & \textbf{51.44} & \textbf{46.00} & \textbf{46.50} \\ 
\hline
\multicolumn{13}{l}{{\cellcolor[rgb]{0.957,0.957,0.957}}\textit{GPT-4V (on a subset of~ test samples)}} \\
Low Resolution & GPT-4 & 512 & 6.00 & 24.00 & 3.57 & 14.41 & 2.12 & \textbf{30.32} & \textbf{56.86} & 41.55 & 45.74 & 34.46 \\
High Resolution & GPT-4 & 768*2000 & 12.50 & \textbf{46.00} & 9.75 & 23.38 & 3.50 & 26.44 & 43.17 & 48.58 & \textbf{60.58} & 37.66 \\
\multicolumn{13}{l}{{\cellcolor[rgb]{0.957,0.957,0.957}}\textit{Ours~(on a subset of~ test samples)}} \\
\textbf{Table-LLaVA} 7B & Vicuna-1.5 7B & 336 & 32.00 & 30.50 & 17.72 & 30.45 & \textbf{18.44} & 29.55 & 40.40 & 51.66 & 40.74 & 50.94 \\
\textbf{Table-LLaVA} 13B & Vicuna-1.5 13B & 336 & \textbf{34.50} & 26.00 & \textbf{18.41} & \textbf{30.54} & 15.88 & 29.87 & 42.88 & \textbf{52.03} & 41.65 & \textbf{51.85} \\
\hline
\end{tabular}
}
\caption{Comparison between GPT-4V and Table-LLaVA on 6 held-in table structure understanding benchmarks. } 
\label{TSU_GPT4V}
\end{table*}

\begin{table*}[t]\footnotesize
\centering
\renewcommand{\arraystretch}{1.2}
\setlength\tabcolsep{3pt}
\scalebox{0.88}{
\begin{tabular}{cccccccccc} 
\hline
\multicolumn{1}{c|}{\multirow{2}{*}{\textbf{Method }}} & \multicolumn{1}{c|}{\textbf{AIT-QA}} & \multicolumn{1}{c|}{\textbf{PubHealthTab}} & \multicolumn{1}{c|}{\textbf{TabMCQ}} & \multicolumn{2}{c|}{\textbf{TSD}} & \multicolumn{1}{c|}{\textbf{TCE}} & \multicolumn{1}{c|}{\textbf{TCL}} & \multicolumn{2}{c}{\textbf{RCE}} \\ 
\cline{2-10}
\multicolumn{1}{c|}{} & \multicolumn{1}{c|}{Acc} & \multicolumn{1}{c|}{Acc} & \multicolumn{1}{c|}{Acc} & \multicolumn{1}{c|}{Row Acc.} & \multicolumn{1}{c|}{Col Acc.} & \multicolumn{1}{c|}{Acc.} & \multicolumn{1}{c|}{Acc.} & \multicolumn{1}{c|}{Row F1.} & Col. F1. \\ 
\hline
\rowcolor[rgb]{0.957,0.957,0.957} \multicolumn{1}{l}{\textit{Previous Best}} & Vary-toy & Monkey & Monkey & LLaVA-1.5 & mPLUG-Owl2 & Monkey & LLaVA-1.5 & Monkey & LLama2+OCR \\
 & \textbf{9.39} & 18.89 & 17.89 & 2.40 & 3.60 & 0.76 & 0.93 & 4.29 & 4.54 \\
\multicolumn{10}{l}{{\cellcolor[rgb]{0.957,0.957,0.957}}\textit{Ours}} \\
Table-LLaVA 7B & 5.48 & \textbf{51.03} & 44.51 & 25.20 & \textbf{16.40} & 11.28 & 26.10 & \textbf{21.97} & 18.14 \\
Table-LLaVA 13B & 6.06 & 48.46 & \textbf{51.51} & \textbf{31.60} & 14.80 & \textbf{11.38} & \textbf{26.17} & 21.94 & \textbf{18.67} \\ 
\hline
\multicolumn{10}{l}{{\cellcolor[rgb]{0.957,0.957,0.957}}\textit{GPT-4V (on a subset of test samples)}} \\
Low Resolution & 19.00 & 59.50 & 66.00 & 8.00 & 15.00 & 10.29 & 17.73 & 27.69 & 50.36 \\
High Resolution & \textbf{62.50} & \textbf{67.00} & \textbf{66.00} & 19.00 & \textbf{38.00} & \textbf{14.36} & \textbf{27.91} & \textbf{48.52} & \textbf{57.14} \\
\multicolumn{10}{l}{{\cellcolor[rgb]{0.957,0.957,0.957}}\textit{Ours (on a subset of test samples)}} \\
Table-LLaVA 7B & 5.00 & 52.50 & 43.50 & 22.00 & 16.00 & 12.73 & 26.27 & 16.57 & 13.91 \\
Table-LLaVA 13B & 6.50 & 53.50 & 45.50 & \textbf{30.00} & 15.00 & 11.92 & 25.45 & 20.77 & 13.78 \\
\hline
\end{tabular}
}
\caption{Evaluation results on 7 held-out tabular benchmarks.}
\label{heldout_benchmark}
\end{table*}

\begin{table*}[t]\footnotesize
\centering
\renewcommand{\arraystretch}{1.2}
\setlength\tabcolsep{3pt}
\begin{tabular}{ccccc} 
\hline
\textbf{OCR Accuracy} & \textbf{TABMWP} & \textbf{TabFact} & \textbf{WTQ} & \textbf{HiTab} \\
Cell-level OCR Accuracy (\%) & 75.35 & 51.48 & 27.09 & 11.05 \\ 
\hline
\textbf{Table Size} &  &  &  &  \\
Ave. Row Number & 6.45 & 14.40 & 26.4 & 23.38 \\
Ave. Col Number & 2.19 & 6.23 & 6.2 & 8.17 \\
Ave. Cell Number (Row*Col) & 14.13 & 83.71 & 163.68 & 191.01 \\ 
\hline
\textbf{Image Resolution (px)} &  &  &  &  \\
Ave. Width*Height & 269*190 & 2354*875 & 1996*1137 & 3194*870 \\ 
\hline\hline
\textbf{Methods} &  &  &  &  \\
Llama2+Oracle & 17.88 & 4.32 & 4.26 & 1.21 \\
Llama2+OCR & 16.35 & 4.21 & 3.91 & 0.77 \\
Gap between Oracle and OCR & 1.53 & 0.11 & 0.35 & 0.44 \\ 
\hline
TableLlama+Oracle & 12.98 & 82.55 & 31.63 & 64.71 \\
TableLlama+OCR & 11.09 & 44.54 & 12.49 & 13.51 \\
Gap between Oracle and OCR & 1.89 & 38.01 & 19.14 & 51.20 \\
\hline
\end{tabular}
\caption{LLM performance on benchmarks with different OCR success rates.}
\label{LLM_performance_with_OCR_success_rates}
\end{table*}

\subsection{Influence of Input Image Resolution}
\label{influence_of_image_resolution}

\begin{figure*}[t]
  \centering
  \includegraphics[width=\linewidth]{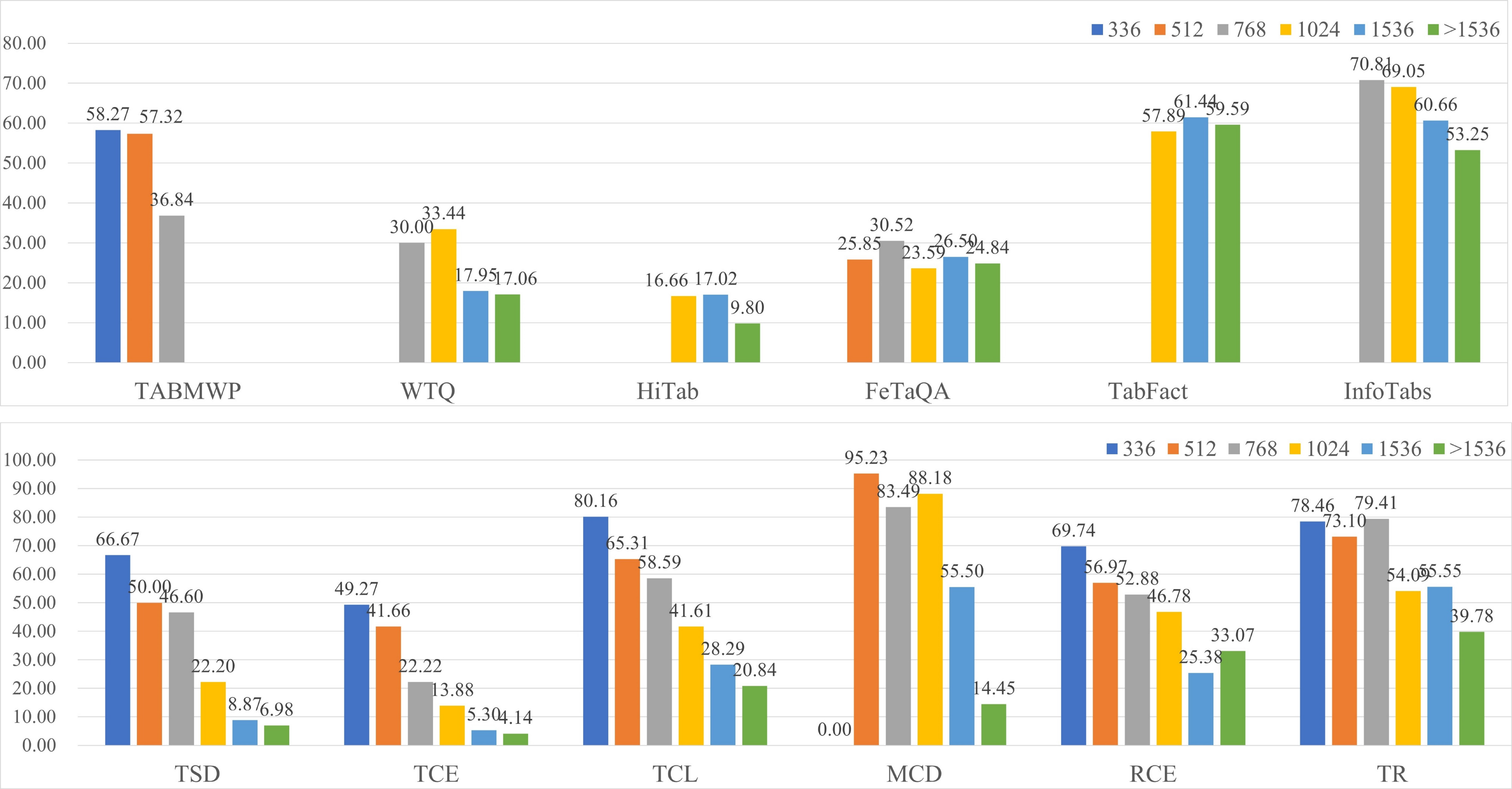}
  \caption{Break-down experimental results of Table-LLaVA 7B by different input image resolution. We divide test samples into 5 groups according to their image resolution, e.g., `512' represents the input image resolution is smaller than 512$\times$512 but larger than 336$\times$336. For TSD, MCD, RCE and TR, we report averaged results.}
  \label{resolution_experiment_results}
  
\end{figure*}

To shed more light on the influence of image resolution on multimodal table understanding, we divide test samples into 5 groups by their image resolution and evaluate model performance on different groups. The results, illustrated in Figure \ref{resolution_experiment_results}, demonstrate that image resolution has a great impact on model performance. The model performance gradually degenerates with the increasing image resolution, which reveals that it is necessary to enlarge the input image solution of MLLMs in order to process extremely large table images.

\subsection{Influence of MMTab on Non-tabular Tasks}
\label{influence_of_MMTab_on_non_tabular_tasks}
We compare Table-LLaVA with its backbone LLaVA-1.5 on two non-tabular benchmarks: TextVQA~\citep{textvqa_benchmark}, a VQA benchmark requiring the understanding of image texts, and LLaVA-Bench (In-the-Wild)~\citep{liu2023llava_1}, a recent general benchmark for MLLMs including 3 task categories (conversation, detail description and complex reasoning). The results are listed in the Table \ref{results_of_table_llava_on_non_tabular_tasks}. Table-LLaVA beats LLaVA-1.5 in most cases under both model sizes, which demonstrates that tabular training data has positive impact on the performance on non-tabular tasks.

\subsection{Influence of OCR Success Rate on LLM Performance}
We compute the cell-level OCR success rates on 4 benchmarks and show the performance of textual LLMs in Table \ref{LLM_performance_with_OCR_success_rates}. As shown in the table, OCR success rates vary a lot among 4 benchmarks, ranging from 11.05\% to 75.35\%. Intuitively, table images with large sizes (i.e. large Ave. Cell Numer) pose greater challenge to OCR engines and thus often lead to low OCR success rates. With OCR success rate decreasing, the performance gap of TableLlama between '+Oracle' and '+OCR' settings significantly increases, which reveals the importance of correct table recognition results. Moreover, compared with TableLlama, the performance gap of Llama2 between two settings is much more lower and less significant, which shows its bottleneck is the ability to understand and follow table-related instructions, rather than OCR results. 

By manually inspecting the OCR results, we find that typical error types include (1) character-level mistakes, e.g., missing the first or last letter, (2) cell-level mistakes, e.g., missing whole cells, mistakenly splitting text in one cell into two cells, very wrong cell text especially for cells with long and intensive text, (3) row or column level mistakes, e.g., missing rows or inserting non-existing rows. (4) structure-level mistakes, e.g., falsely recognizing a merged cell as a non-merged cell or vice versa.

\subsection{Case Study}
\label{case_study}
\begin{figure*}[t]
  \centering
  \includegraphics[width=0.95\linewidth]{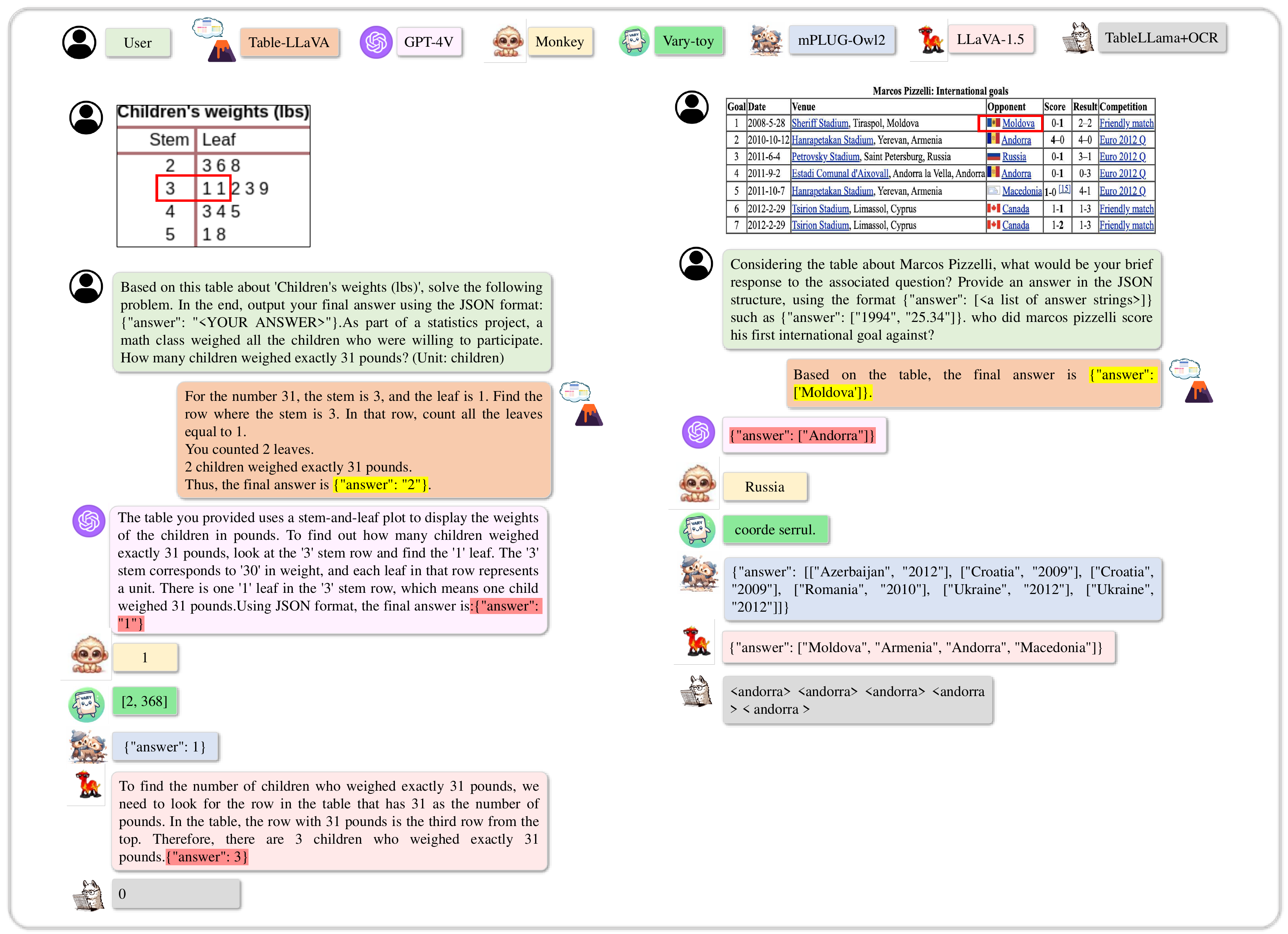}
  \caption{Case study on the TABMWP and WTQ benchmark. For the TABMWP benchmark, the model needs to conduct multi-step reasoning to obtain the final answer. 
  }
  \label{case_study_1}
  
\end{figure*}

\begin{figure*}[t]
  \centering
  \includegraphics[width=0.95\linewidth]{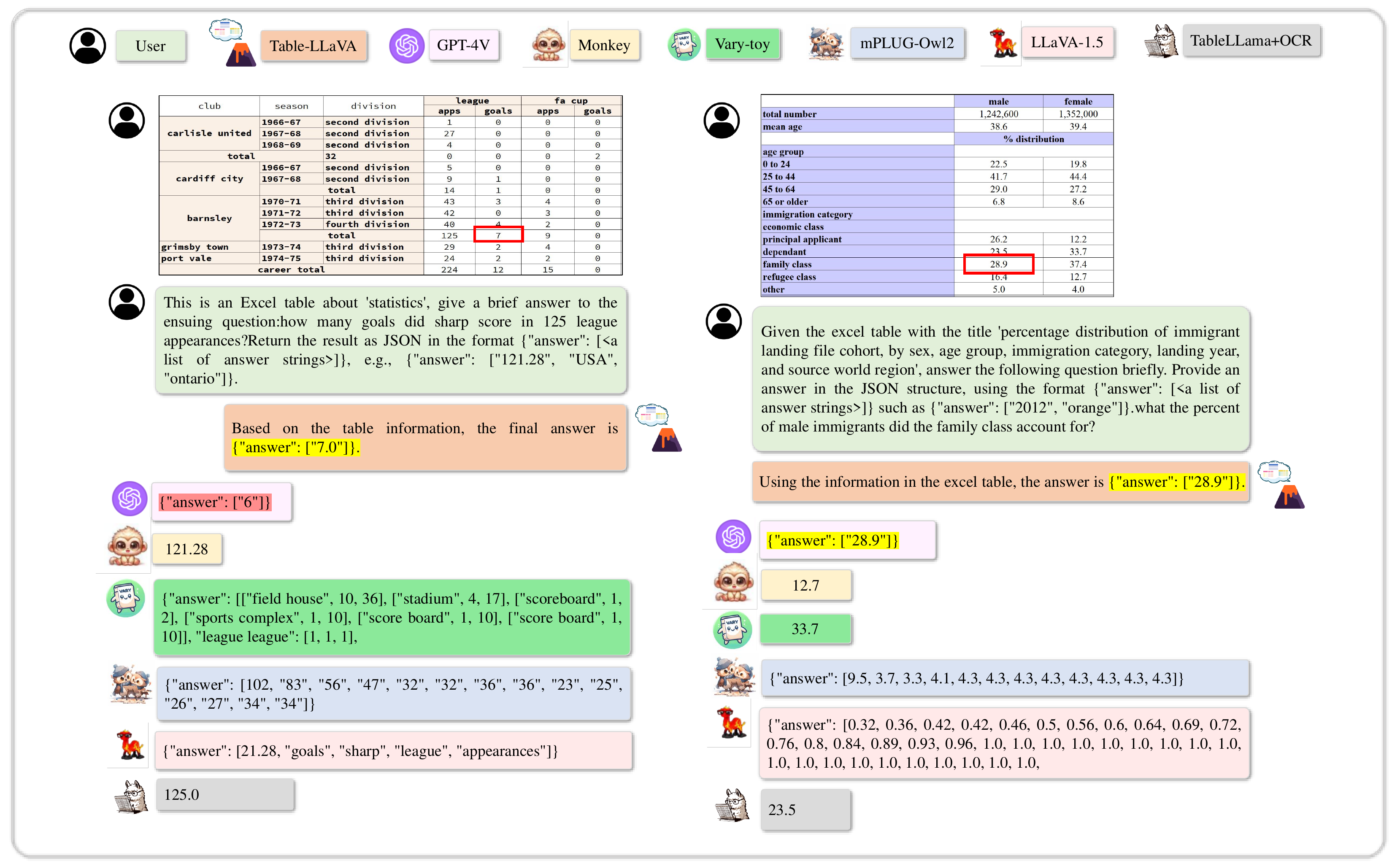}
  \caption{Case study on the HiTab benchmark, where the model is required to comprehend hierarchical tables with merged cells. }
  \label{case_study_2}
  
\end{figure*}

\begin{figure*}[t]
  \centering
  \includegraphics[width=0.95\linewidth]{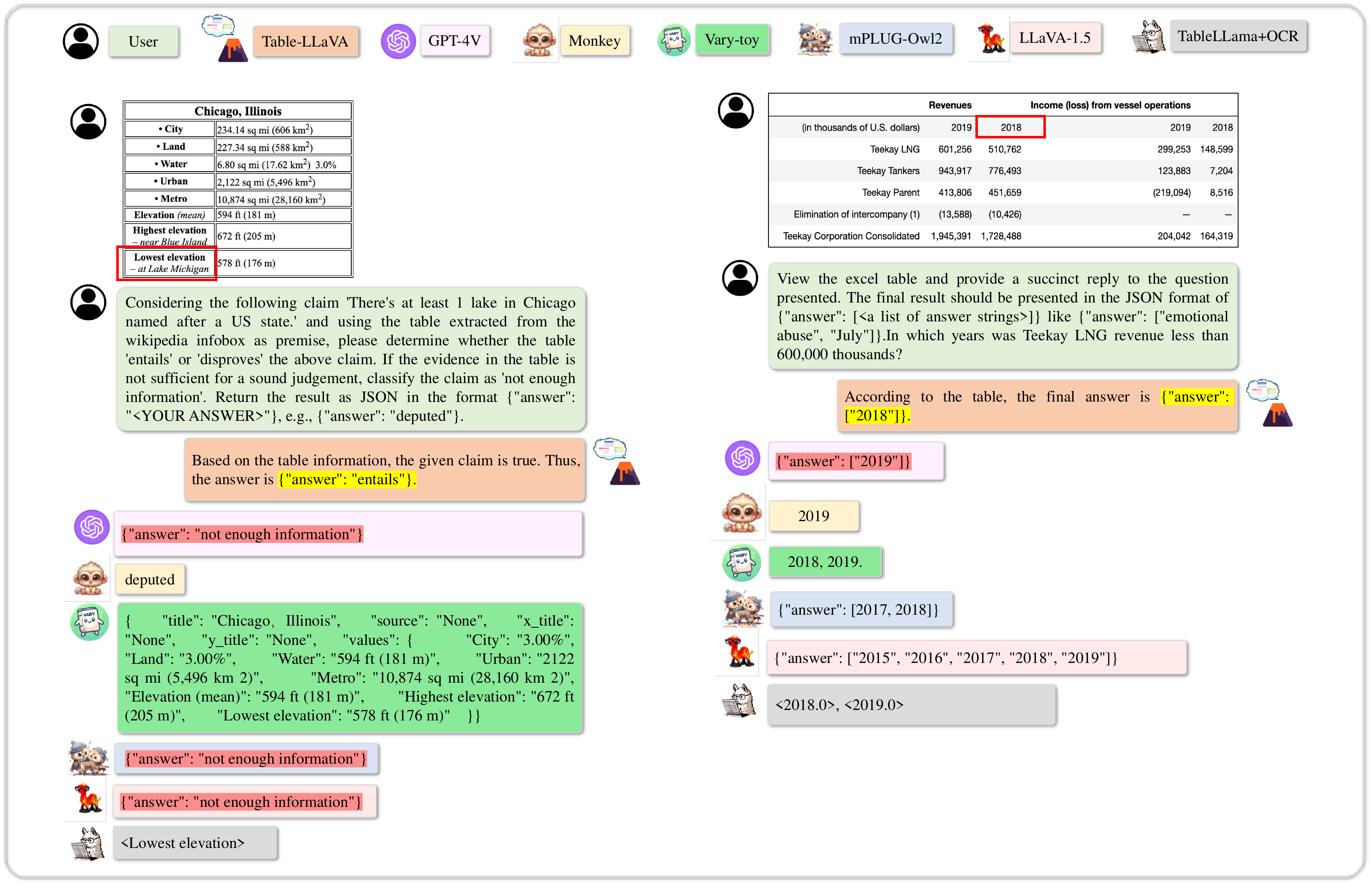}
  \caption{Case study on the InfoTab and TAT-QA benchmarks. Common table structures usually have the first row as column header, but table structures in IntoTab are quite different, where the first column contains row headers. Diverse table structures in MMTab pose unique challenge to existing MLLMs. }
  \label{case_study_3}
  
\end{figure*}

\begin{figure*}[t]
  \centering
  \includegraphics[width=0.95\linewidth]{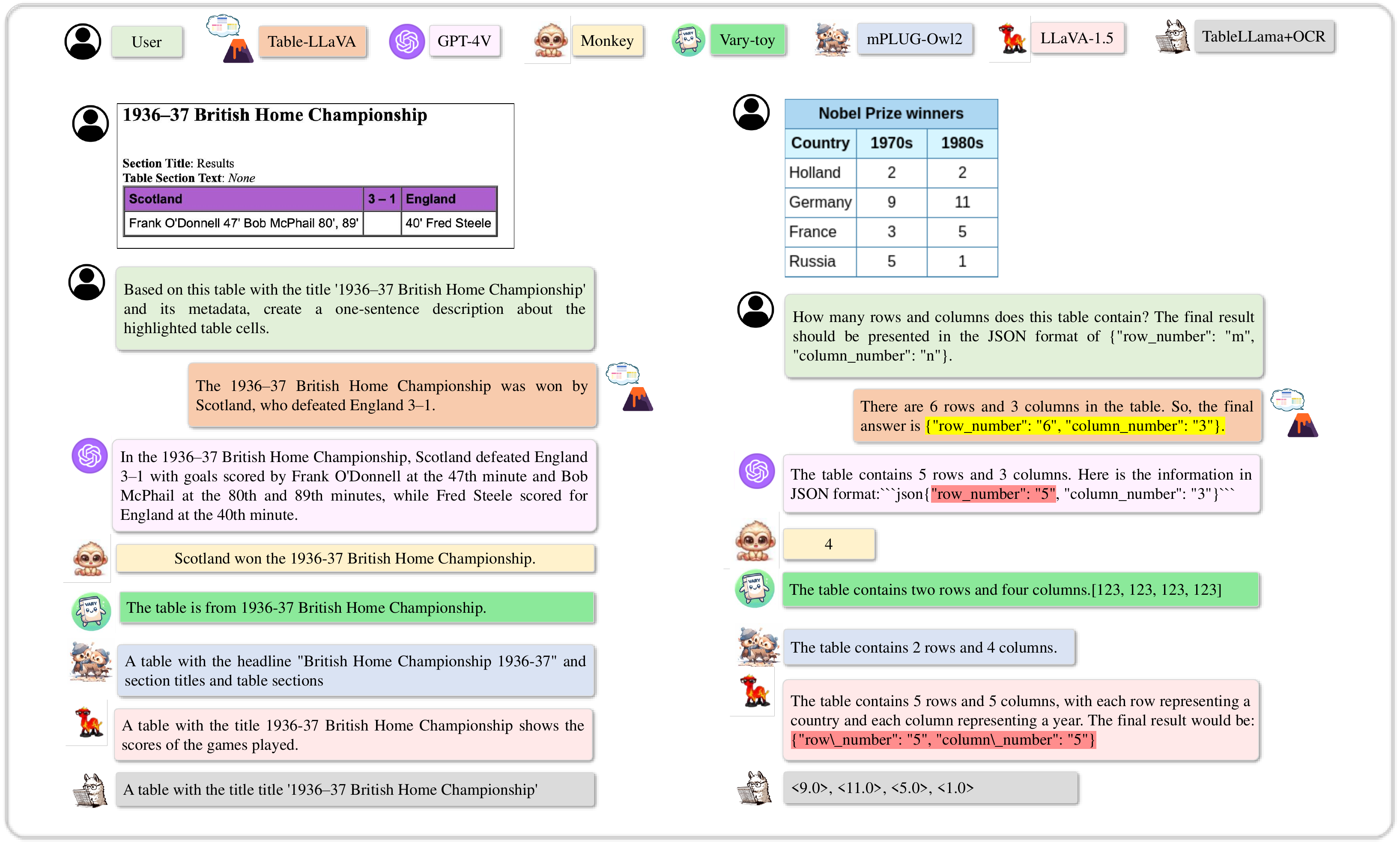}
  \caption{Case study on the ToTTo and TSD benchmarks. Though facing a relatively small and simple table, existing powerful MLLMs may fail to determine the row number and column number of the table. The basic ability to understand diverse table structures has been overlooked by previous MLLM study and the proposed MMTab alleviates this problem. 
  }
  \label{case_study_4}
  
\end{figure*}

\begin{figure*}[t]
  \centering
  \includegraphics[width=0.95\linewidth]{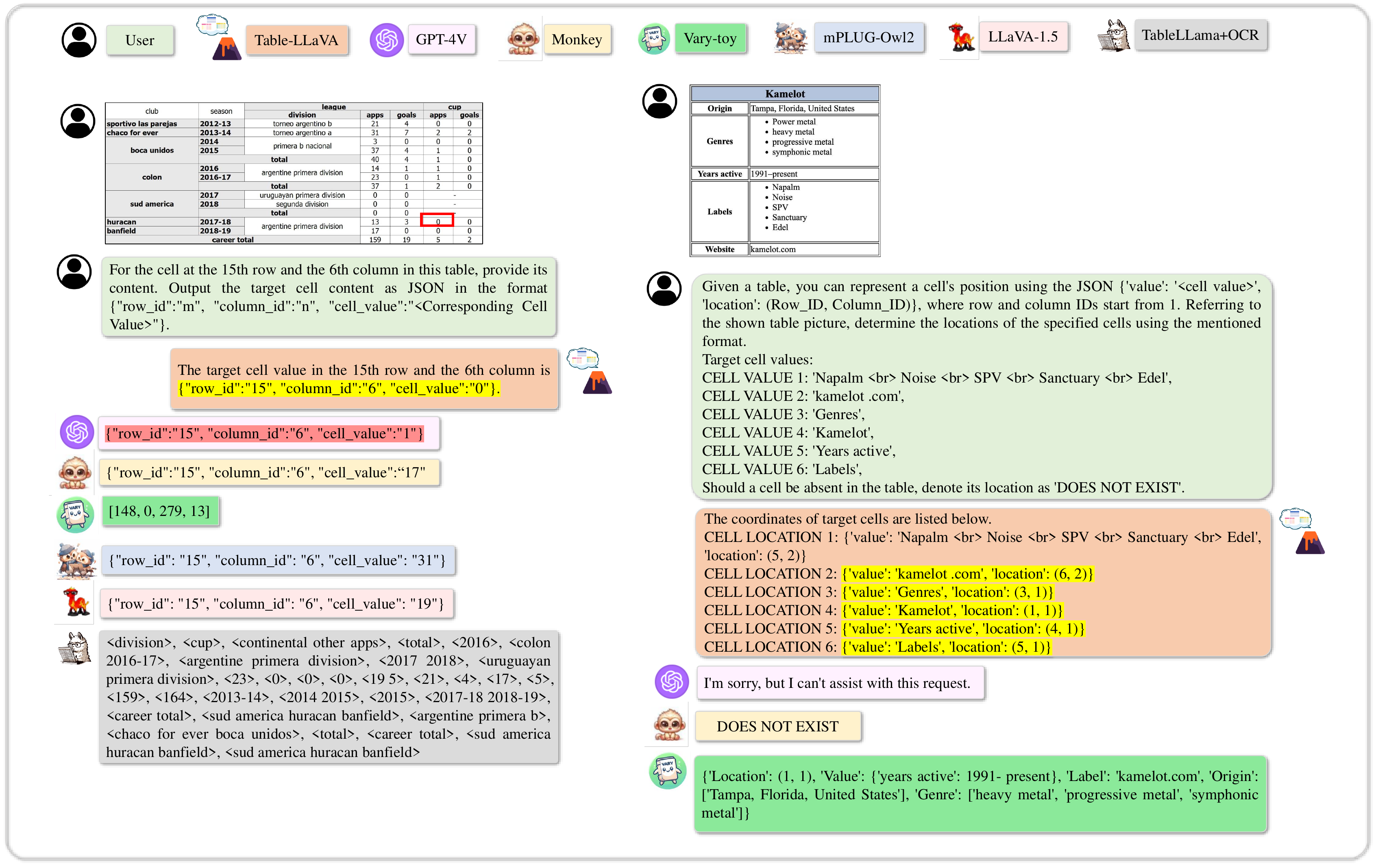}
  \caption{Case study on the TCE and TCL benchmarks, where the model is required to extract the target cell content or find the target cell location based on the table image. This task is trivial for human readers yet is challenging for existing MLLMs, which reveals the gap between current MLLMs and the human-level table understanding ability.
  }
  \label{case_study_5}
  
\end{figure*}

\begin{figure*}[t]
  \centering
  \includegraphics[width=0.9\linewidth]{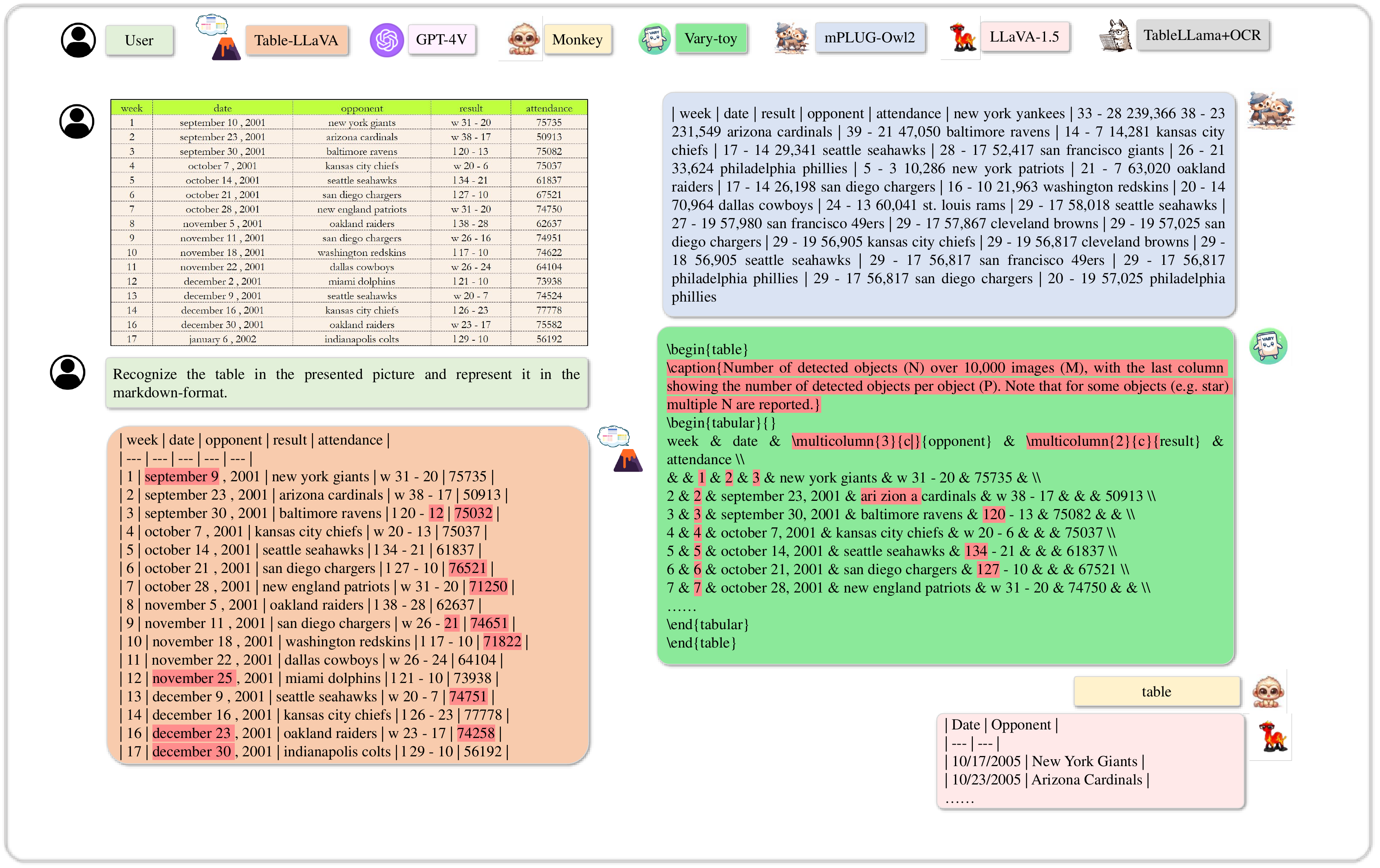}
  \caption{Case study on the TR (table recognition) benchmark. Though Table-LLaVA generate the markdown sequence with correct table structure, its OCR ability is restricted by the input resolution and needs to be enlarged.  In the future work, the proposed MMTab dataset can be combined with MLLMs with stronger OCR ability and larger input image resolution like Monkey and LLaVA-Next to build more powerful tabular MLLMs.
  }
  \label{case_study_6}
\end{figure*}

\begin{figure*}[t]
  \centering
  \includegraphics[width=0.9\linewidth]{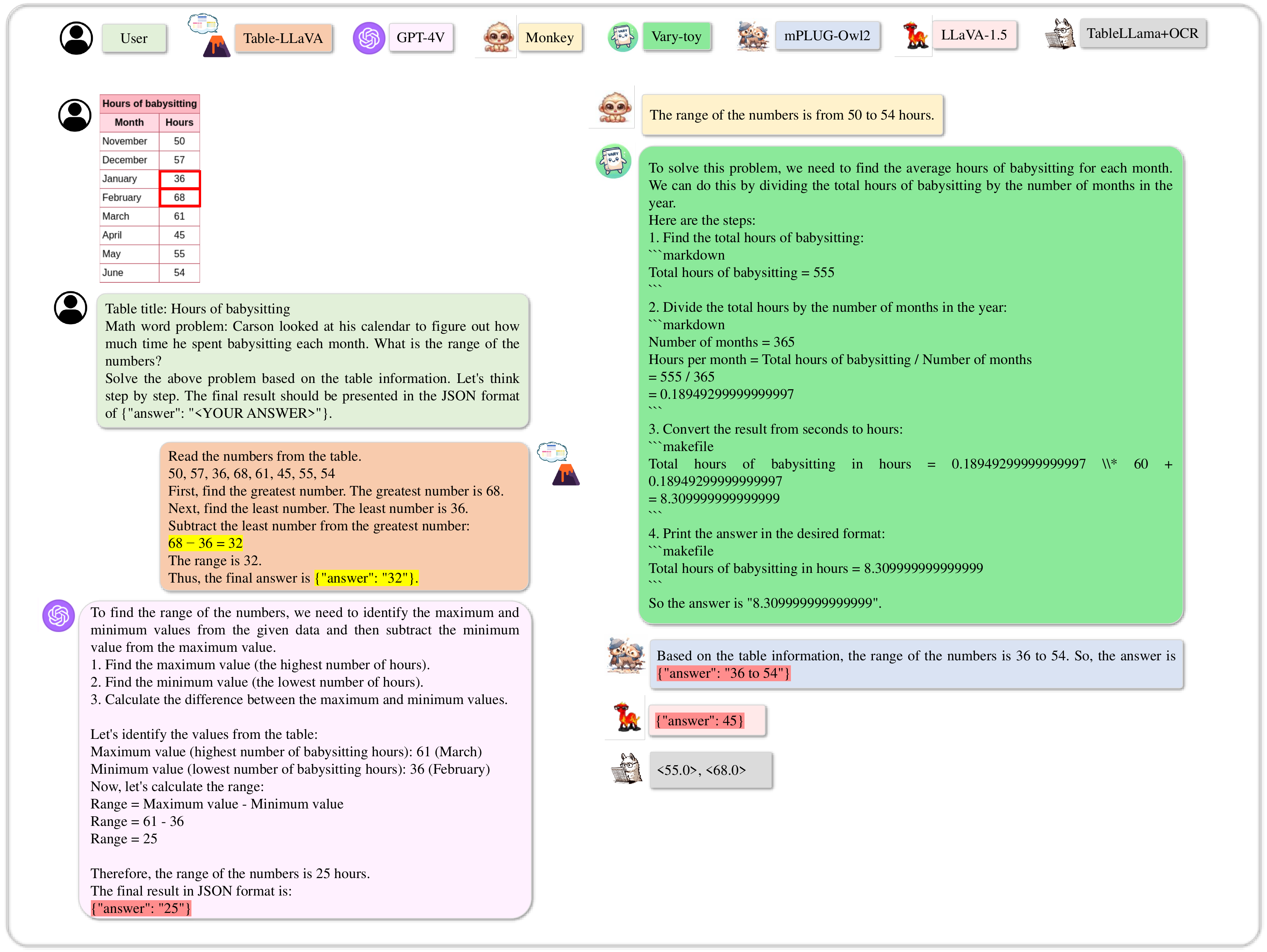}
  \caption{Case study on the TABMWP benchmark. In this case, the model needs to conduct table-based mathematical reasoning such as finding the largest number in the table or do math computations. Moreover, more external tools like Python Interpreter~\citep{chen2023program} could be integrated with Table-LLaVA to build MLLM-based table agents.}
  \label{case_study_7}

\end{figure*}

We conduct a side-by-side qualitative analysis to compare Table-LLaVA with other (M)LLMs on different benchmarks, as illustrated in Figure \ref{case_study_1}-\ref{case_study_7}. The results demonstrate that Table-LLaVA can handle a series of table tasks and possesses better multimodal table understanding ability than existing open-source MLLMs. For instance, as can be seen in Figure \ref{case_study_1}, Table-LLaVA provides both the intermediate reasoning steps and the correct final answer for the math word problem based on table image, whereas other MLLMs including GPT-4V fail to give the correct answer. The proposed MMTab dataset can be directly utilized in the training process of future MLLMs to boost their multimodal table  understanding ability.


\end{document}